\documentclass[10pt,twocolumn,letterpaper]{article}
\usepackage[pagenumbers]{cvpr}
\newcommand{\papertitle}{RegHarness: Regulating What Code-as-Policy Agents Keep from Steps to Skills}%

\usepackage{tabularx}
\usepackage{multirow}
\usepackage{microtype}

\usepackage{pifont}
\newcommand{\cmark}{\ding{51}}
\newcommand{\xmark}{\ding{55}}
\usepackage{algorithm}
\usepackage{algpseudocode}
\usepackage{cuted}
\usepackage{currfile}
\AtBeginDocument{%
  \crefname{algorithm}{Alg.}{Algs.}%
  \Crefname{algorithm}{Algorithm}{Algorithms}%
}

\newcommand{\pending}[1]{{\color{red}[#1]}}

\renewcommand{\paragraph}[1]{\vspace{.5em}\noindent\textbf{#1.}}

\newcommand{\sysname}{RegHarness\xspace}
\newcommand{\rtgov}{run-time governance\xspace}
\newcommand{\brref}{between-run refinement\xspace}
\newcommand{\devagent}{Claude Opus~5.5\xspace}
\newcommand{\llm}{\texttt{deepseek-flash}\xspace}

\newcommand{\rats}{RATs\xspace}
\newcommand{\cfgA}{\mbox{RATs\,+\,stepwise replay}\xspace}
\newcommand{\cfgO}{\mbox{RATs\,+\,refined skills}\xspace}
\newcommand{\cfgH}{\mbox{+\,Harness}\xspace}

\newcommand{\libero}{LIBERO-PRO\xspace}
\newcommand{\molmo}{MolmoSpaces\xspace}

\newcommand{\auditor}{auditor\xspace}
\newcommand{\controller}{controller\xspace}
\newcommand{\resetreplay}{reset-and-replay\xspace}

\newcommand{\planaudit}{plan audit\xspace}
\newcommand{\codeaudit}{code audit\xspace}
\newcommand{\stepaudit}{step audit\xspace}
\newcommand{\finalaudit}{final audit\xspace}
\newcommand{\gateone}{\ding{202}}
\newcommand{\gatetwo}{\ding{203}}
\newcommand{\gatethree}{\ding{204}}
\newcommand{\gatefour}{\ding{205}}

\newcommand{\dec}[1]{\textsf{\small #1}}
\newcommand{\decs}[1]{\textsf{\footnotesize #1}}
\newcommand{\dcommit}{\dec{commit}\xspace}
\newcommand{\dretry}{\dec{retry}\xspace}
\newcommand{\dreplan}{\dec{replan}\xspace}
\newcommand{\dabort}{\dec{abort}\xspace}

\newcommand{\taskcheck}{task check\xspace}
\newcommand{\steptest}{step test\xspace}
\newcommand{\committed}{committed steps\xspace}

\newcommand{\pub}{$^{\dagger}$}

\usepackage{flushend,needspace}
\usepackage[hidelinks]{hyperref}
\newcommand{\mainref}[1]{\cref{#1}}

\makeatletter
\patchcmd{\@maketitle}{\vskip .375in}{\vskip 3pt}{}{}
\patchcmd{\@maketitle}{\vspace*{24pt}}{\vspace*{10pt}}{}{}
\patchcmd{\@maketitle}{\vspace*{12pt}}{}{}{}
\apptocmd{\@maketitle}{\noindent\begin{minipage}{\textwidth}
\centering
\includegraphics[width=\linewidth]{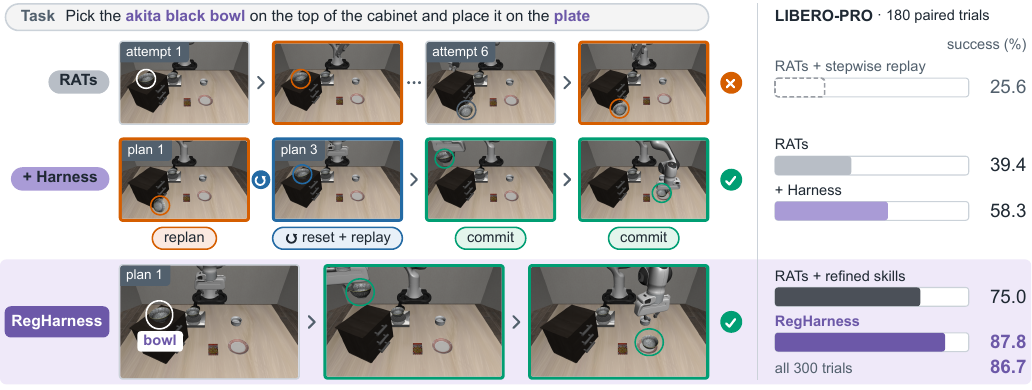}
\captionsetup{hypcap=false}
\captionof{figure}{\textbf{\rats fails this trial in all six attempts; \rtgov and \sysname both solve it.}
\rats~\cite{zhang2026rats}, run with our low-cost model, writes six programs; none lifts the bowl.
With the released skills, \rtgov (\cfgH) replans after a grasp knocks the bowl off the cabinet and succeeds on its third plan; \sysname succeeds on its first.
Blue marks a reset scene with the \committed replayed; orange borders mark rejected executions (for \rats, failed \taskcheck{}s), green ones commits, and row-end marks show the \taskcheck.
Right: success of all five configurations on the 180 paired trials, with gray bars for the \rats loop, purple bars for \rtgov and dark bars for the refined skills, and of \sysname on its 300.
}%
\captionsetup{hypcap=true}
\label{fig:\currfilebase}%
\end{minipage}\par
\par\vspace{6pt}}{}{}
\makeatother
\title{\papertitle}
\author{%
  Shengming Liu\textsuperscript{1,*}, Yunyi Gao\textsuperscript{1,*}, Xinyi Wang\textsuperscript{1,*}, Jianyong Sun\textsuperscript{2}, Shuai Wang\textsuperscript{1},\\
  Yi Deng\textsuperscript{1}, Li Chenye\textsuperscript{1}, Mingcai Zhou\textsuperscript{3,\textdagger}, and Sim Kuan Goh\textsuperscript{1,\textdagger}\\[3pt]
  {\small\textsuperscript{1}School of Artificial Intelligence and Robotics, Xiamen University Malaysia, Sepang, Selangor, Malaysia}\\
  {\small\textsuperscript{2}Embodied AI, Beijing Zhongke Huiling Robot Technology Co., Ltd., Beijing, China}\\
  {\small\textsuperscript{3}Institute of Automation, Chinese Academy of Sciences, Beijing, China}%
}
\hypersetup{pdfauthor={Shengming Liu, Yunyi Gao, Xinyi Wang, Jianyong Sun, Shuai Wang, Yi Deng, Li Chenye, Mingcai Zhou, Sim Kuan Goh}}

\newcommand{\authornotes}{%
  \begingroup
  \renewcommand{\thefootnote}{\fnsymbol{footnote}}%
  \footnotetext[1]{These authors contributed equally.}%
  \footnotetext[2]{Corresponding authors.}%
  \endgroup
}

\date{}
\hypersetup{pdftitle={\papertitle}}
\begin{document}
\flushbottom
\maketitle
\authornotes
\begin{abstract}
Code-as-Policy agents generate, execute, and revise robot programs, but often lack explicit mechanisms to govern what persists across executions: which actions are retained within a trial and which code modifications carry over to subsequent runs. This leaves two critical aspects of agent behavior insufficiently controlled: step-level execution and cross-run adaptation.
We present \sysname, a two-level harness that governs both forms of persistence for the \rats agent team, without relying on a vision-language-action (VLA) policy or demonstrations. Within each trial, \rtgov evaluates each proposed step from a reset scene, replaying the \committed, while a fixed-rule \controller commits the step only when both \rats' verifier and a task-long LLM \auditor approve it. Between runs, a coding agent uses logged failures to develop and test revisions to skills, primitives, and rules, which remain disabled by default until explicitly enabled by the authors.
Using a single low-cost model across all run-time roles, \sysname achieves an 87.8\% success rate (158/180) on paired \libero trials, compared with 39.4\% for \rats, and a 72.0\% success rate on \molmo trials. A 2$\times$2 ablation study of \sysname's two levels shows that its skill refinement alone increases \rats' successful trials from 71 to 135, whereas \rtgov alone achieves 105 and \rats' native stepwise \resetreplay mode achieves only 46.
These findings suggest that, for Code-as-Policy agents operating with low-cost models, task success depends not only on how generated programs are executed, but also on the quality of the reusable skills they invoke and how those skills evolve across runs.
\end{abstract}

\section{Introduction}
\label{sec:intro}

Consider the task in \cref{fig:teaser}: pick the black bowl on top of a cabinet and place it on a plate.
\rats~\cite{zhang2026rats}, a Code-as-Policy~\cite{liang2022cap} agent team that we run with a low-cost language model, makes six attempts at this \libero~\cite{zhou2025liberopro} task, each a program for the whole plan.
Its static checker rejects one program before it runs, and none of the five that run lifts the bowl: the first leaves it on the cabinet, the last two knock it onto the table.

Over the 180 paired \libero trials, \rats solves 39.4\% (71/180) within six attempts but only 13 on the first.
In its published loop, \rats judges each step only after the whole program has run; a retry, though prompted to keep what worked, is a complete program run again from the reset scene, so a passed step can change and must pass again.
All 692 retries are complete programs, and 86\% (493/574) of those that follow a diagnosed failure change code outside the failed step (criteria in the supplementary material).
Nor does a clean verdict mean the work is done: in 117 of the 670 failed executions, the verifier flags no failed step.

What these agents lack is a rule for what to keep: which executed steps stay committed within a trial, and which code changes carry over to the next run.
Testing one step at a time does not supply it by itself: the released \rats code already contains a stepwise mode, off by default and not described in its paper, that resets the scene and replays the passed steps before every try, but it commits a step on the verifier's pass alone and, once a step's tries run out, commits the step unverified so that the plan can go on.
A reset protects a retried step from what the failed test left behind, but it also erases what that test revealed; a trial therefore needs a memory that outlives the reset, and a commit rule that never keeps an unverified step instead of letting the coder's next program decide which code survives.
Between runs, the code itself must change: even after we add such a commit rule to \rats (\cfgH, \cref{sec:ablation}), 75 of the 180 trials still fail, 30 of them at a grasp that never lifts a non-flat object (19) or at a placement off target (11).

\begin{figure*}[t]
  \centering
  \IfFileExists{fig/fig2_overview.pdf}{%
    \includegraphics[width=\textwidth]{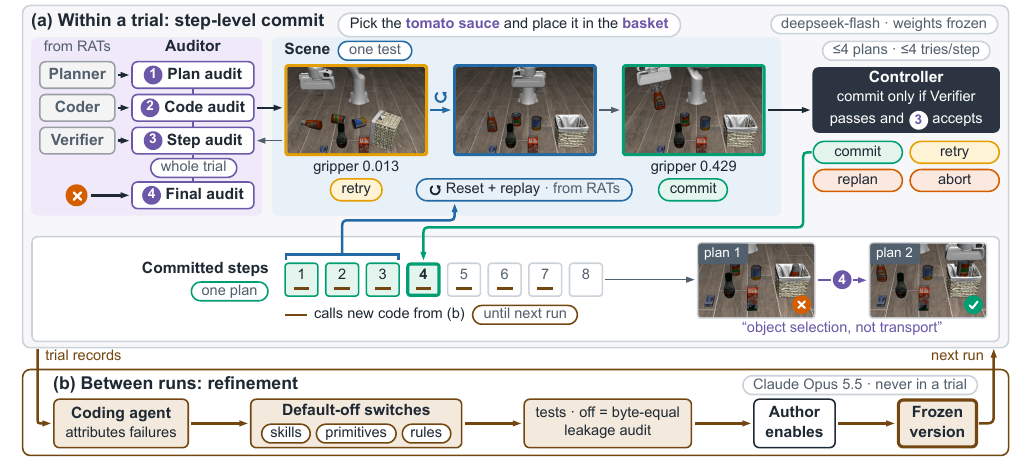}%
  }{%
    \fbox{\parbox[c][2.75in][c]{0.96\textwidth}{\centering\pending{Figure 2 (two-level overview) from the architecture group: figures/v2\_20261114/arch\_final/fig.pdf}}}%
  }
  \caption{\textbf{\sysname regulates what persists within a trial and between runs.}
  (a)~The \auditor audits plans~\gateone, drafts~\gatetwo, \steptest{}s~\gatethree{} and finished plans that fail the \taskcheck~\gatefour.
  Each \steptest resets the scene and replays the \committed on the executor of \rats' stepwise mode.
  The \controller commits a step only if the verifier passes it and the \auditor proposes \decs{commit}.
  (b)~The coding agent (\devagent) turns failures into default-off switches that the authors enable.}%
  \label{fig:overview}
\end{figure*}

\sysname regulates what persists at these two time scales (\cref{fig:overview}).
Within a trial, it wraps \rats' planner, coder and verifier and reuses the executor of the stepwise mode, so every \steptest resets the scene and replays the \committed (\resetreplay); one LLM \auditor per trial audits plans, code, \steptest{}s and finished plans that fail the benchmark's \taskcheck, and a \controller of six fixed rules turns its proposal and the remaining budget into \dcommit, \dretry, \dreplan or \dabort, never committing a step that the verifier has not passed.
Between runs, a coding agent (\devagent in Claude Code) reads full trial logs and frames and turns each fix to skills, primitives or rules into a default-off switch that must pass tests, and a byte-for-byte equivalence check when off, before the authors can enable it.
Reset-and-replay, which needs a scene that software can reset, is confined to simulation.

Coding agents and optimizers already revise robot harnesses and skills between runs~\cite{elmaaroufi2026rho,lu2026aspire,wei2026scale,darwin2026harnesspai,zhang2026vista,deng2026dynaharness}, most of them behind an automatic acceptance test that our author-gated loop lacks, and the concurrent RegenHarness separates a model's proposal from a deterministic commit gate on a quadruped~\cite{wang2026regenharness}.
\sysname is a new harness framework that improves \rats in two ways, without a VLA policy or demonstrations: within a trial it replaces \rats' loop with a commit rule over steps tested from a reset scene, and between runs it refines the skill library that \rats shares across tasks.

With one low-cost model in every role, \sysname solves 158 of the 180 \libero trials on initial states 0--2 (87.8\%), where \rats solves 71 (39.4\%), and 260 of 300 (86.7\%) over five initial states; on \molmo it solves 72.0\% (144/200).
The 2$\times$2 splits this gain unevenly between \sysname's two levels: its refined skills add 64 trials under \rats' loop and 53 under \rtgov, whereas replacing \rats' loop with \rtgov adds 34 trials with the released skills and 23 with the refined ones (exact McNemar $p=1.2\times10^{-4}$ and $2.9\times10^{-4}$).
\rats' own stepwise mode solves 46, fewer than \rats itself, so the reset alone does not explain the margin of \rtgov.
A \sysname trial costs an estimated CNY~2.16 (about US\$0.32 at CNY~6.67 per US\$) in model calls, against 1.51 (US\$0.23) for \rats (\cref{sec:failures}).
Both levels were debugged on initial state~0 only and tested directly on states 1--4 (\cref{sec:setup}).

Our contributions are:
\begin{itemize}
  \item \textbf{Rule-decided commits over steps tested from a reset scene, paired with author-gated refinement of a shared skill library}, around an existing Code-as-Policy agent and without a VLA policy or demonstrations (\cref{sec:method}).
  \item \textbf{A 2$\times$2 over the control loop and the skill library on the same 180 paired \libero trials}, with \rats' own stepwise mode as a control (\cref{sec:main,sec:ablation}).
  \item \textbf{An account of what happens inside trials, what still fails and what it costs}, failures caused by \rtgov included (\cref{sec:metrics,sec:failures,sec:molmo}).
\end{itemize}

\section{Related work}
\label{sec:related}

\paragraph{Code-as-Policy agents}
Language-model agents write robot programs against perception and control APIs~\cite{liang2022cap,huang2023voxposer,singh2023progprompt}
and revise them from execution feedback, one program at a time.
Voyager commits a program to its skill library only after a self-verification module confirms the task~\cite{wang2023voyager}.
CaP-X never resets the environment within a trial, so each new program starts in the scene the previous one left~\cite{fu2026capx}.
\rats instead reruns a complete program from a reset scene and judges its steps after the run, whereas \sysname makes the step the unit that is tested and committed (\cref{sec:setting}).

\paragraph{Run-time verification and commit gates}
The harnesses closest to \rtgov shape what the model perceives~\cite{li2026k1,zhang2026waa,chu2026openrua}
or check outcomes during execution~\cite{zhou2024monitor,galanti2026pigey,grant2026spark,gong2026tinyastral,wang2026auriga,darwin2026harnesspai},
and recover in whatever scene a failure leaves; Recova first restores that scene with recovery skills~\cite{liu2026recova}.
The concurrent RegenHarness goes furthest in separating proposal from commitment: its models only propose, and a deterministic gate advances a quadruped's trusted task state only after checking the verifier's verdict~\cite{wang2026regenharness}, but the paper does not describe undoing a rejected execution.
We instead test each step from a reset scene, as RE-0 restores simulator snapshots to check code corrections before distilling them~\cite{zhang2026re0}, but inside the evaluated trial:
a rejected test is discarded by the next reset, and passed steps persist as code that is replayed (\cref{sec:replay}).

\paragraph{Between-run refinement by coding agents}
ASPIRE has Claude Code repair programs and admit validated repairs to a skill library~\cite{lu2026aspire}, SHAPER has the frozen planner revise its own skills and context code~\cite{wang2026shaper}, and HarnessPAI already separates two timescales, revising across rollouts a fixed program that checks each step within one~\cite{darwin2026harnesspai}.
Concurrent systems bring this loop to \libero around a frozen $\pi_{0.5}$ policy~\cite{black2025pi05}, driven by a Codex meta-agent that explores failures in VISTA~\cite{zhang2026vista}, by failure attribution to harness layers in DynaHarness~\cite{deng2026dynaharness} and by a self-generated curriculum in RoboACE~\cite{zhao2026roboace}; EmbodiedEvo instead lets the acting agent write its own code skills~\cite{li2026embodiedevo}.
RHO's training-time search over policy repositories~\cite{elmaaroufi2026rho}, Scale and Selection~\cite{wei2026scale}, TinyAstral~\cite{gong2026tinyastral}, VISTA, DynaHarness and RoboACE keep a change only after a fixed comparison with its predecessor, and the first three, like ASPIRE, score the result on held-out seeds or tasks.
Our \brref lacks the first safeguard, since the authors choose which tested, default-off switches to enable; it draws only on initial state~0, and states 1--4 are held out (\cref{sec:refine,sec:setup}).

\paragraph{Concurrent work and our position}
Without a VLA policy or demonstrations, AMP1 has a frontier model write code against a motion interface and solves most \libero episodes in one attempt~\cite{chen2026amp}, and RURL trains a reasoning model to act with no harness at all~\cite{tang2026rurl}.
On what each paper reports, none of the concurrent systems above commits steps tested from a reset scene, shares one skill library across tasks without a VLA policy and splits its gain between two levels on one set of paired trials: RegenHarness gates commits without a reset, AMP1 writes code afresh per episode and AMP1+ adds an agent-written library but no commit decisions, and those that refine code between runs report component ablations at most (comparison table in the supplementary material).
Their tasks, budgets, models and success rules differ from ours, so their published rates (\cref{tab:published}) serve only as context.

\section{Method}
\label{sec:method}

\sysname reuses \rats' planner, coder and verifier~\cite{zhang2026rats} and adds an \auditor, which can rewrite plans or drafts before they run but after a \steptest only proposes, and a \controller, which decides; between runs it refines skills (\cref{fig:overview}).

\subsection{Setting and notation}
\label{sec:setting}
A trial runs one task instruction $g$ from one saved initial state $s_0$ with $P=4$ plans and $J=4$ tries per step.
The $p$-th plan $\pi^{(p)}=(\sigma_1,\dots,\sigma_{n_p})$ lists steps, the $j$-th try of step~$k$ is a draft $\tilde c_k^{(j)}$ of its code, and a draft that runs is executed once, in a \steptest that resets the simulator and replays the code $C_{k-1}=(c_1,\dots,c_{k-1})$ of the \committed:
\begin{equation}
  \bigl(s_k^{(j)},e_k^{(j)}\bigr)=\mathrm{Exec}\bigl(\mathrm{reset}(s_0),\;c_1\oplus\cdots\oplus c_{k-1}\oplus\tilde c_k^{(j)}\bigr),
  \label{eq:steptest}
\end{equation}
where $\oplus$ joins step code into one program after the skill preamble and $e_k^{(j)}$ is the execution record, for which $\mathrm{ok}(e)$ holds when the reset succeeds and the program ends without an exception, a timeout or an uncertain change to the scene.
The verifier returns $v\in\{\textsf{succeeded},\textsf{failed},\textsf{ambiguous}\}$, and the benchmark's \taskcheck $T(s)\in\{0,1\}$ is read by the \controller and shown to no model while a plan runs.

\sysname keeps \rats' planner, coder (now asked for one step at a time), static checker and Per-Step Verifier, all on the \auditor's language model and with the skills frozen within a trial.
It drops \rats' outer retry loop, Planner Verifier, rehearsal (\cref{sec:setup}), feedback generator and SubAgent, and adds planning rules, read by the planner and the \planaudit, that \rats does not have.

Run-time governance reuses the executor of the stepwise mode of the released \rats code (\texttt{-{}-multiturn-reset}, \cref{sec:intro}) but not the mode's commit loop.

\subsection{Testing a step from a reset scene}
\label{sec:replay}
A draft can be judged only by running it, and a failed run has already changed the scene: in the task of \cref{fig:teaser}, a grasp try in the first plan of \rtgov with the released skills knocks the bowl off the cabinet.
Every \steptest therefore starts from $s_0$ with the \committed replayed (\cref{eq:steptest}), so the next reset discards what a rejected test did, and the second plan's first step finds the bowl back on the cabinet.
Replay re-executes code instead of restoring a saved state and is not checked against the run it repeats, although a replayed step reuses the grasp sample it drew when committed.
Like \rats' own reset between attempts, \resetreplay needs a scene that can be reset, and even there a cabinet settles differently after each reset in two \libero tasks (scene comparisons in the supplementary material).

Testing every try from a reset scene costs re-execution, not language-model calls: a trial of $N$ \steptest{}s, the $t$-th testing step $k_t$ after replaying $k_t-1$ committed steps, executes $B$ steps of code,
\begin{equation}
  B=\sum_{t=1}^{N}k_t,\qquad \text{replayed share}=\frac{B-N}{B},
  \label{eq:cost}
\end{equation}
and each \steptest adds a reset, perception on the replayed steps, a verifier call and a \stepaudit call.

\subsection{A task-long auditor}
\label{sec:auditor}
The reset also erases what the failed test showed, unless something outlives it: the scene lasts one \steptest, the \committed one plan, the \auditor's context the whole trial, and the code until the next run, which only \brref changes (lifetime labels in \cref{fig:overview}).
The \auditor is one resumed Codex CLI conversation per trial, on the language model of the other roles, so what a failed try revealed reaches the next draft and plan after its scene is gone.
It cannot call tools, cannot reach the environment and answers each audit in a fixed JSON schema (prompts and schemas in the supplementary material).

The \planaudit~\gateone{} reads each plan with the task, a text summary of the scene and the API and skill interfaces, and the \codeaudit~\gatetwo{} reads each draft with the static checker's findings; neither sees an image.
Both accept, rewrite or reject: a rewrite runs in place of the original, and a rejection, like a block by the static checker, uses up the plan or the try.
After each \steptest the \stepaudit~\gatethree{} reads the executed code, an excerpt of its output, the verifier's verdict, the remaining budget and the first and last camera frames, and proposes \dcommit, \dretry, \dreplan or \dabort with a correction (\cref{fig:qualitative}).
Apart from what the code prints, no audit receives the reward, the completion flag, depth, point clouds, masks or simulator state; the \finalaudit~\gatefour{} is told that the task is not complete.
Since the \auditor is the same model that writes the code, its rewrites can introduce errors of their own (\cref{sec:failures}).

\begin{table*}[t]
  \caption{\textbf{\sysname solves 158 of the 180 paired \libero trials, against 71 for \rats and 135 for \rats with the refined skills, and 86.7\% (260/300) over five initial states.} All rows use \llm in every role, score one run per task and initial state and count successes (of 30 per cell; of 50 in the last row); initial state~0 is the development state and states 1--4 are test states, and \rats counts scored executions only (\cref{sec:setup}). \emph{Step commit}: --, none; forced, a step whose tries run out is kept unverified.}
  \label{tab:main}
  \centering
  \small
  \setlength{\tabcolsep}{3.6pt}%
  \begin{tabular}{@{}lllr*{6}{c}r@{}}
    \toprule
    & \multicolumn{2}{c}{Setting} & & \multicolumn{2}{c}{Object} & \multicolumn{2}{c}{Goal} & \multicolumn{2}{c}{Spatial} & \\
    \cmidrule(lr){2-3}\cmidrule(lr){5-6}\cmidrule(lr){7-8}\cmidrule(lr){9-10}
    Method & Skills & Step commit & Trials & Pos & Task & Pos & Task & Pos & Task & All \\
    \midrule
    \rats & released & -- & 180 & 18 & 12 & 12 & 12 & 9 & 8 & 71/180 (39.4) \\  %
    \cfgA & released & verifier, forced & 180 & 11 & 15 & 7 & 8 & 1 & 4 & 46/180 (25.6) \\  %
    \cfgO & refined & -- & 180 & 25 & 24 & 21 & 19 & 20 & 26 & 135/180 (75.0) \\  %
    \cfgH & released & verifier + auditor & 180 & 21 & 15 & 18 & 15 & 19 & 17 & 105/180 (58.3) \\  %
    \sysname & refined & verifier + auditor & 180 & 27 & 26 & 26 & 23 & 27 & 29 & \textbf{158/180 (87.8)} \\  %
    \midrule
    \sysname, initial states 0--4 & refined & verifier + auditor & 300 & 43 & 42 & 44 & 38 & 45 & 48 & \textbf{260/300 (86.7)} \\  %
    \bottomrule
  \end{tabular}

  \vspace{4pt}
  \caption{\textbf{Published \libero success rates (\%), for context.} Rows marked \pub{} use other models, trial counts and protocols and are not a controlled comparison with \cref{tab:main}; RHO\pub{} evolves on unperturbed LIBERO tasks and HarnessPAI\pub{} on 15 of its 50 evaluation seeds.}
  \label{tab:published}
  \footnotesize
  \setlength{\tabcolsep}{3.6pt}%
  \begin{tabular}{@{}ll*{6}{c}r@{}}
    \toprule
    Method & Setting & Obj.\ Pos & Obj.\ Task & Goal Pos & Goal Task & Spa.\ Pos & Spa.\ Task & All \\
    \midrule
    CaP-Agent0\pub~\cite{fu2026capx} & rerun in~\cite{zhang2026rats} & 27.0 & 31.0 & 29.0 & 16.0 & 13.0 & 23.0 & 23.2 \\  %
    RATs\pub~\cite{zhang2026rats} & Play skills; model not stated & 61.0 & 63.0 & 43.0 & 36.0 & 29.0 & 31.0 & 43.8 \\  %
    SPARK\pub~\cite{grant2026spark} & Qwen3.8-27B & 86.4 & 82.4 & 71.6 & 61.8 & 65.4 & 83.0 & 75.1 \\  %
    OpenRUA\pub~\cite{chu2026openrua} & Claude Opus 5 & 98.0 & 97.0 & 79.0 & 79.0 & 98.0 & 100.0 & 91.8 \\  %
    RHO\pub~\cite{elmaaroufi2026rho} & no LLM at test time & 12.2 & 37.2 & 50.6 & 55.6 & 53.0 & 61.6 & 45.0 \\  %
    HarnessPAI\pub~\cite{darwin2026harnesspai} & drives $\pi_{0.5}$-LIBERO & 98.6 & 99.2 & 94.6 & 97.8 & 93.8 & 95.2 & 96.5 \\  %
    \bottomrule
  \end{tabular}
\end{table*}

\subsection{A controller of fixed rules}
\label{sec:controller}
The \controller turns every proposal into a decision by fixed rules, so the same proposal, checks and budget always lead to the same action.
Step~$k$ is committed only when
\begin{equation}
  \begin{gathered}
  \kappa=\mathrm{ok}(e)\wedge\bigl(v=\textsf{succeeded}\bigr),\\
  \text{commit}\iff\kappa\wedge\bigl(a=\textsf{commit}\bigr),
  \end{gathered}
  \label{eq:commit}
\end{equation}
where $a$ is the \auditor's proposal.
One deterministic check can overturn $v$: the \emph{holding check} fails a step that should grasp or hold an object but ends with the gripper commanded closed and a reading below 0.03, if the step has a try left; the coding agent set this threshold from gripper readings in earlier runs on initial state~0, so it is tuned on the benchmark and can misfire on a grasp that holds.

Six rules, applied in order, map $a$, $\kappa$ and the remaining budget to a decision, and a \taskcheck that holds ends the trial as a success before any rule applies (rules and pseudocode in the supplementary material).
Whatever the \auditor proposes, no step is committed unless the verifier passes it, even when its tries run out, and only the budget or the \taskcheck ends a trial; in practice the rules rarely override the \auditor (\cref{sec:metrics}), so they add this guarantee rather than decisions of their own.

\subsection{When a finished plan fails the task check}
\label{sec:finalaudit}
A plan whose steps are all committed can still fail the \taskcheck, because each step is verified only against its own goal; when it does and a plan is left, the \finalaudit~\gatefour{} reads the verdicts and the last output and names the step and the measurement that let the error through.
It lists at most five changes, which open the next plan, but it cannot change the outcome or request an execution.
In the trial of \cref{fig:overview}, the first plan, finding no ``tomato sauce'', commits all eight steps with a bottle; the audit traces the failure to ``object selection, not transport'', and the second plan takes the can.

\subsection{Between-run refinement}
\label{sec:refine}
Run-time governance decides which code is committed, but what that code can do depends on the skills, primitives and rules it calls, which change only between runs.
In \brref, sub-agents of the coding agent, \devagent in Claude Code, read the full log and frames of every failed trial in a batch, name the earliest step and mechanism that failed, and turn each refinement into a default-off switch in a new copy of the code.
A switch must pass unit tests, be byte-for-byte equivalent to the previous version when off, survive an adversarial check by further agents and pass a leakage audit, which from the fourth round on forbids reading object positions or other data directly from the simulator in anything the run-time models see or execute, the \taskcheck excepted (leakage checks in the supplementary material).
The authors choose which switches to enable before the code is frozen and evaluated, in five rounds on \libero; in the third they enabled 41 of 55 proposed switches.

Of the 194 changes registered against the released code, 78 make it run consistently with our model and scoring, 49 implement \rtgov itself, 40 revise skills and primitives, and 27 are undecided (change register in the supplementary material); run-time governance with the released skills keeps the first two groups and three undecided items.
The skill revisions, such as grasp checks that require fingers stopped on an object, camera-frame point clouds for the grasp network and numbered look-alike instances, grow the library from 59 to 71 entries.

This loop is gated by the authors and lacks the automatic acceptance test of \cref{sec:related}; the coding agent also reads success checks, full logs and, on \molmo during design, the benchmark's ground truth, of which the run-time models receive only the \finalaudit's notice that the task is not complete.

\section{Experiments}
\label{sec:experiments}

\subsection{Setup}
\label{sec:setup}
\paragraph{Benchmarks}
\libero~\cite{zhou2025liberopro} supplies the 60 tasks used by CaP-X~\cite{fu2026capx} and \rats~\cite{zhang2026rats}, ten from each of the Object, Goal and Spatial suites of LIBERO~\cite{liu2023libero} under position and ten under task perturbations.
We use the stored initial states $s=0,\dots,4$ of each task, and each task--state pair is one trial, scored once per system, and voided runs were rerun in full, none after a success.

\paragraph{Compared systems}
For \rats we rerun the released code and skills, with the changes needed to run them, at the code's defaults of six attempts per trial and 180\,s per execution, which its paper does not state.
\rats with its built-in stepwise reset-and-replay (\cfgA) switches on \texttt{-{}-multiturn-reset} in the same code (\cref{sec:setting}), with four attempts, four tries per step, no rehearsal and a two-line fix to the step index its per-step verifier receives.
\rats with the refined skills (\cfgO), \sysname's between-run level alone, keeps the default loop and carries 51 of the 64 skill-level refinements, the rest belonging to run-time governance or never reaching the model.
Run-time governance with the released skills (\cfgH), the within-trial level alone, uses 5,844 characters of planning rules, four plans of four tries per step and 480\,s per execution; \sysname adds the refined skills and extends the rules to 10,380 characters.
All five systems run initial states 0--2, the 180 paired trials, on which one trial is worth 0.56 points, and \sysname also runs states 3 and 4; the supplementary material lists every difference between the systems.

\paragraph{Metric and protocol}
Success is the benchmark's \taskcheck under the \rats protocol~\cite{zhang2026rats}: the agent retries within a trial until the check passes or the budget is spent, and only the environment decides.
The released code also rehearses each program from the reset scene before its scored execution, checking only for errors, a step its paper does not describe; a rehearsal that ends with the \taskcheck true, as in 65 of 180 \rats trials, does not count, since the released loop discards it and runs the scored execution (rehearsal details in the supplementary material).
All five systems read the \taskcheck during execution, as do the main SPARK results~\cite{grant2026spark}, and we ran no control without it; under run-time governance an error raised after the check has passed does not undo a success (3 of 105 successes for \cfgH, 4 of 260 for \sysname), whereas the \rats paper counts every policy crash as a failure.
The per-step verifier that all five systems run is meant to judge a step from its goal, code, printed values and frames, but the released code cannot locate a step's code, mostly when the step's description contains a closing parenthesis, and the verifier then judges from frames alone, in 7.7\% of \rats' per-step verdicts and 45.8\% of \sysname's; \cfgA runs it with a two-line fix to the step index and the harness systems with their own input format, so its evidence differs between systems in a way whose effect we did not measure, although only the \taskcheck decides success.

\paragraph{Development exposure and model}
We debugged both levels on initial state~0 only and ran initial states 1--4 as test states; which changes \cfgH drops was decided after the results of \sysname and \rats were known.
State~0 is thus a development state and a third of the paired trials, so the gain of the refined skills on these trials is an upper estimate, and \cfgO inherits this exposure through its skills.
Every role in every trial uses the provider alias \llm at high thinking effort.

\begin{figure*}[t]
  \centering
  \includegraphics[width=\textwidth]{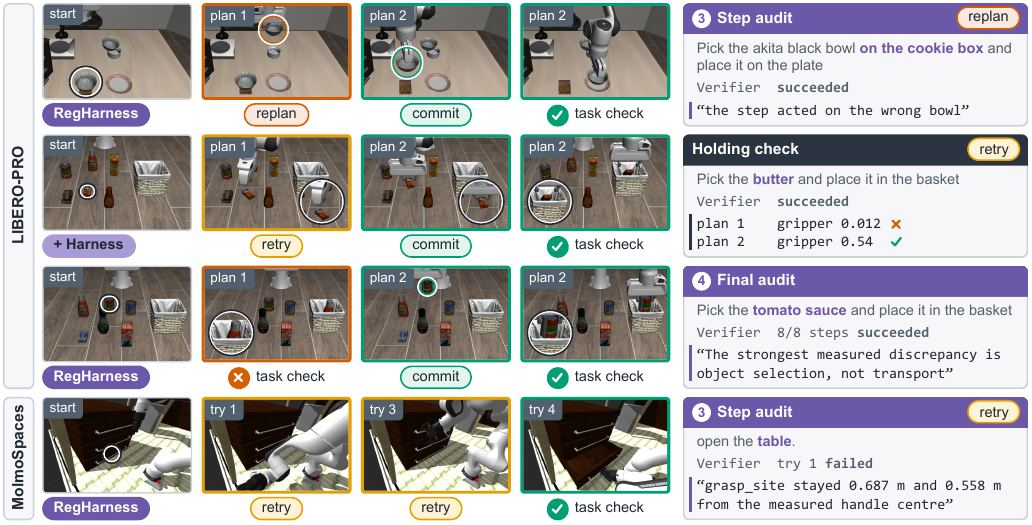}
  \caption{\textbf{Selected successes in which run-time governance stops an execution that the verifier accepts or redirects one that it rejects.}
  Each row is one trial, labelled with its configuration; borders mark stopped (amber, red) and corrected (green) executions, and the other pills are \controller decisions.
  In rows~1--3 (\libero), a \stepaudit~\gatethree{} rejects a wrong-bowl grasp, the holding check overturns an accepted empty grasp before a \stepaudit retry, and a \finalaudit~\gatefour{} traces a failed \taskcheck to the wrong object.
  In row~4 (\molmo), the \stepaudit turns three tries the verifier rejects into retries.}%
  \label{fig:qualitative}
\end{figure*}

\subsection{Main results on \libero}
\label{sec:main}
\paragraph{\sysname solves 86.7\% of \libero trials with one low-cost model}
Over five initial states \sysname solves 260 of 300 trials (\cref{tab:main}; 95\% CI $[79.7,\,93.0]$, bootstrap over the 60 tasks), with 54, 53, 51, 53 and 49 of 60 from states 0 to 4.
State~0, whose failures drove all refinements, is highest, but a paired test over the 60 tasks does not separate the five states (Cochran's $Q$, $p=0.32$); on the test states 1 and 2, \sysname solves 104 of 120 and \rats 49.
\sysname solves 158 (87.8\%) of the 180 paired trials and keeps all but 3 of the 71 that \rats solves.
The published rates in \cref{tab:published} give context, not a controlled comparison, and we do not rerun RHO or HarnessPAI, whose evolved \libero solvers are not released.

\subsection{Where the gain comes from}
\label{sec:ablation}
\begin{figure*}[t]
  \centering
  \includegraphics[width=\textwidth]{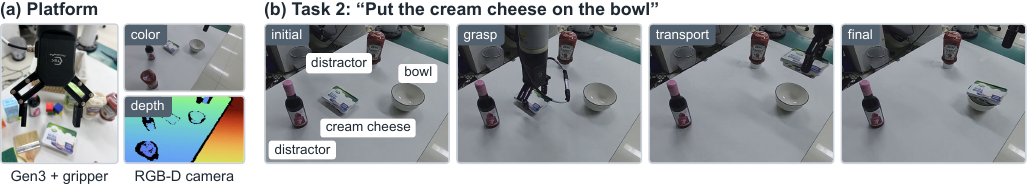}%
  \caption{\textbf{Real-robot setup and an example task.}
  (a)~The Kinova Gen3 arm with its two-finger gripper, and color and depth images from the single fixed RGB-D camera.
  (b)~Task~2, ``put the cream cheese on the bowl'': the initial scene, the grasp, the transport and the final state.}%
  \label{fig:realrobot}
\end{figure*}
\begin{figure}[t]
  \centering
  \includegraphics[width=\columnwidth]{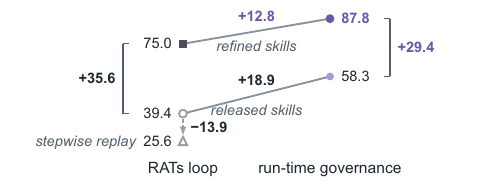}
  \caption{\textbf{Both levels of \sysname raise success, each by less when the other is present, while \rats' own stepwise replay lowers success.}
  Success (\%) on the 180 paired \libero trials for each pairing of skill library (lines) and control loop (columns), and for \cfgA (triangle); slopes give the change from \rtgov and brackets the change from the refined skills, in points.}%
  \label{fig:attribution}
\end{figure}

\paragraph{Both levels of \sysname raise success, and between-run refinement adds the larger share}
Replacing the released skills with the refined ones (\cref{fig:attribution}) raises success from 71 to 135 trials in \rats' loop and from 105 to 158 under \rtgov ($+35.6$ and $+29.4$ points; exact McNemar $p=3.3\times10^{-13}$ and $4.9\times10^{-13}$; 95\% CIs, bootstrapped over tasks, in the supplementary material).
Replacing \rats' loop with \rtgov raises it from 71 to 105 with the released skills and from 135 to 158 with the refined ones ($+18.9$ and $+12.8$; $p=1.2\times10^{-4}$ and $2.9\times10^{-4}$).
Each level gains 11 trials less when the other is present, so on these trials the two levels overlap in part; the skill gains are upper estimates (\cref{sec:setup}), and the one under \rtgov also contains the longer planning rules and two further sentences of audit instructions.

These compare whole configurations: against \rats' loop, \rtgov has four plans instead of six attempts, 480 instead of 180\,s per execution, its own prompts, planning rules and consistency fixes, and it reads the \taskcheck after 17.7 \steptest{}s per trial where \rats reads it after 4.1 scored executions. Since a \rats program's rehearsal and scored execution end differently in 76 of 741 cases, single flips mean little.

\paragraph{\rats' own stepwise reset-and-replay solves 46 of 180 trials}
With the skills, code and model of \rats and the executor that \rtgov reuses, \cfgA gains 18 trials and loses 43 against \rats ($-13.9$ points; $p=0.0019$), and \rats solves 56 within its first four attempts, the budget of \cfgA.
Against \rtgov with the released skills it gains 10 and loses 69 ($p=5.5\times10^{-12}$).
Its logs show where the mode departs from \cref{eq:commit}: of the 4,181 steps it commits, 1,924 (46.0\%) are committed unverified after their four tries, and 243 of its 10,699 step executions (2.3\%) reach the 180\,s limit that the replayed steps share with the step under test.
Testing steps from a reset scene therefore does not by itself explain the margin of \rtgov; since \cfgA also lacks the planning rules, the holding check, our consistency fixes and the longer time limit, the 59 trials between the two are not owed to the \auditor and \controller alone.

\subsection{Inside a trial}
\label{sec:metrics}
\paragraph{Rejected tests and rewrites are routine, changed proposals and \finalaudit{}s rare}
We did not ablate the \auditor's context, the replay or the \controller's rules, so the counts below show how often each mechanism acts, not what it contributes (full counts, replay failures and false commits in the supplementary material).
Over its 300 trials, \sysname rejects 1,119 of 4,440 \steptest{}s, 452 of them after the arm has moved, and 87 of its 260 successes contain at least one of these 452.
The \auditor rewrites 67.3\% of the plans and 48.5\% of the drafts before they run, and after a \steptest it withholds \dcommit from 135 of the 3,400 that the verifier passes (row~1 of \cref{fig:qualitative}) but never proposes \dcommit for a step the verifier has not passed.
The \controller changes only 62 of 4,440 proposals, all because no plan is left; the \finalaudit runs in 25 trials, 16 of which a later plan solves (row~3); and in 56 of the 260 successes the \taskcheck ends the trial after a \steptest that the \auditor did not propose to commit.
On the paired trials a \sysname \steptest replays 3.1 \committed on average, so 75.8\% of executed step segments are replays (\cref{eq:cost}).

\subsection{What still fails, and the cost}
\label{sec:failures}
\paragraph{What still fails is object identity, flat plates and a low handle}
An LLM agent assigns each failed trial the first error that no later plan overcomes (procedure and per-trial classes in the supplementary material).
Of \sysname's 40 failures over five initial states, 20 come from the wrong object or part (16) or a misread instruction or target region (4), and 16 from a flat plate not grasped or pushed (11, in three tasks) or a drawer handle too low to reach (5, in one task); 4 have other causes.
Of the 75 failures of \rtgov alone, 5 are its own: two follow a rewrite by the \auditor (low confidence), and in three the holding check, which catches the empty grasp in row~2 of \cref{fig:qualitative}, overturns a grasp that had lifted the bowl.
Of \sysname's 40 failures, we checked 13 and found none caused by \rtgov; two of the other 27 may be.

\paragraph{The system is not free}
On the paired trials a \sysname trial costs an estimated CNY~2.16 (US\$0.32), 65.6\% of it for the \auditor, against 1.51 (US\$0.23) for \rats, and takes 66 minutes on average against 50.
The refined skills make a success cheaper with either loop, CNY~1.38 (US\$0.21) for \cfgO against 3.83 (US\$0.57) for \rats and 2.46 (US\$0.37) for \sysname against 3.85 (US\$0.58) for \cfgH, while run-time governance makes a trial dearer with either library, CNY~2.25 (US\$0.34) against 1.51 with the released skills and 2.16 against 1.04 (US\$0.16) with the refined ones (cost estimation in the supplementary material).
\cfgA is the most expensive, CNY~3.24 (US\$0.49) per trial and 12.68 (US\$1.90) per success, since every try of every step is written and judged on its own.

\subsection{\molmo}
\label{sec:molmo}
\paragraph{\sysname solves 72.0\% (144/200) of \molmo trials}
The 40-entry skill library that \rats released for \molmo~\cite{kim2026molmospaces} stays byte-identical, while the grasp and perception functions it calls, the interface documentation and the rules were refined against the 25 failures of a first port that solved 15 of 40; all 200 trials, five repeats of those scenes with 26 to 30 solved per repeat, are therefore in-sample (95\% CI $[60.0,\,83.0]$, bootstrap over scenes).
Six pick-and-place instructions name no placement target and count as 30 failures, and in 12 successes the program raised an error after the \taskcheck had passed; the \rats paper reports 38.0\%\pub{} on them (task-level results in the supplementary material).

\subsection{Real robot}
\label{sec:realrobot}
We ported \rats and \sysname to a 7-DoF Kinova Gen3 arm with one fixed RGB-D camera (\cref{fig:realrobot}a) and use three tasks: (1)~put a ketchup bottle in a basket, (2)~put a cream-cheese package on a bowl, and (3)~swap a cube on a platform with the cube below it.
With 15 trials per task and system (\cref{tab:realrobot}; protocol, and how \sysname runs without a software reset, in the supplementary material), \sysname solves 28 of 45 trials against 21 for \rats, a difference not significant at this sample size (Fisher's exact test, $p=0.20$).
\par\vspace{4pt}
\noindent\begin{minipage}{\columnwidth}
  \centering
  \captionof{table}{\textbf{Real-robot successes} out of 15 trials per task.}
  \label{tab:realrobot}
  \small
  \begin{tabular}{@{}lcccc@{}}
    \toprule
    Task & 1 & 2 & 3 & All \\
    \midrule
    \rats & 8 & 7 & 6 & 21/45 \\
    \sysname & 12 & 9 & 7 & 28/45 \\
    \bottomrule
  \end{tabular}
\end{minipage}
\par

\section{Conclusion}
\label{sec:conclusion}

On the 180 paired \libero trials, \sysname's refined skills alone lift \rats from 71 to 135 successful trials, its \rtgov alone achieves 105, and their combination reaches 158, compared with 46 for \rats' own stepwise replay. These results highlight the importance of skill quality in Code-as-Policy agents, particularly when operating with low-cost models. While testing steps from reset scenes provides additional benefits, its contribution is most evident when integrated with the broader \rtgov framework.
In our experiments, both levels were debugged on initial state~0 and evaluated directly on states 1--4. Since state~0 accounts for a third of the paired trials, the observed gain from skill refinement may be an upper estimate.
We also evaluated \sysname on a Kinova Gen3 arm, where it solves 28 of 45 trials, compared with 21 for \rats. Although preliminary, this result provides an initial indication of its potential beyond simulation.
Reset-and-replay is currently limited to simulation. As a next step, we will evaluate the frozen systems on initial states 5--9 and investigate replacing reset-based execution on the robot with a recovery skill~\cite{liu2026recova}.

\flushcolsend
\clearpage
\raggedcolsend
\appendix
\section*{Appendix}
\renewcommand{\theHfigure}{appendix.\arabic{figure}}
\renewcommand{\theHtable}{appendix.\arabic{table}}
\providecommand{\theHalgorithm}{\arabic{algorithm}}
\renewcommand{\theHalgorithm}{appendix.\arabic{algorithm}}
\makeatletter
\setlength{\@fptop}{0pt}
\setlength{\@dblfptop}{0pt}
\makeatother
\setcounter{dbltopnumber}{2}
\suppressfloats[t]%
\setlength{\dbltextfloatsep}{12pt plus 2pt minus 4pt}%
\setlength{\textfloatsep}{12pt plus 2pt minus 4pt}

\setcounter{figure}{0}
\setcounter{table}{0}
\setcounter{algorithm}{0}
\renewcommand{\thefigure}{S\arabic{figure}}
\renewcommand{\thetable}{S\arabic{table}}
\renewcommand{\thealgorithm}{S\arabic{algorithm}}

\noindent \Cref{sec:supp_gov,sec:supp_refine} detail \sysname's two levels, \cref{sec:supp_config,sec:supp_pertask} the five \libero systems and the comparisons behind the attribution, and \cref{sec:supp_failures,sec:supp_molmo} failures, cost, \molmo and the real robot; \cref{sec:supp_concurrent} compares concurrent work.
Unless stated otherwise, \libero numbers refer to the 180 paired trials on initial states 0--2.

\section{Run-time governance}
\label{sec:supp_gov}
\label{sec:supp_prompts}

\paragraph{Auditor}
The \auditor is one Codex CLI conversation per trial, backed by \llm at thinking effort ``high''; each audit is one turn, whose reply must validate against that audit's JSON schema (\cref{tab:supp_audits}). The client compacts the conversation automatically at 900,000 tokens, so ``task-long'' means the same conversation, not a verbatim record.

\begin{table}[t]
  \centering
  \caption{\textbf{The four audits.} All are turns of one \auditor conversation per trial; only the \stepaudit receives images.}
  \label{tab:supp_audits}
  \scriptsize
  \renewcommand{\arraystretch}{1.1}
  \setlength{\tabcolsep}{2.5pt}
  \begin{tabularx}{\columnwidth}{@{}>{\raggedright\arraybackslash}p{0.85cm}>{\raggedright\arraybackslash}X>{\raggedright\arraybackslash}p{2.65cm}@{}}
    \toprule
    Audit & Shown & JSON reply \\
    \midrule
    \gateone{} Plan & task, scene summary, API documentation, the plan (skills by interface only), the planning rules & \texttt{decision} (accept, revise, reject), \texttt{rationale}, \texttt{defects}, \texttt{output\_plan} \\
    \gatetwo{} Code & task, API documentation, the step and its skill records, the draft, the static checker's report & \texttt{decision}, \texttt{rationale}, \texttt{defects}, \texttt{final\_body} \\
    \gatethree{} Step & the step, executed code, test output (first 1,000 and last 3,000 characters), verifier status and reason, tries and plans left, first and last frames & \texttt{action} (commit, retry\_step, replan, abort), \texttt{rationale}, \texttt{corrective\_action}, \texttt{edit\_scale} \\
    \gatefour{} Final & that the \taskcheck is false; committed steps' verdicts; end of the last output; last gripper reading & \texttt{process\_consistent}, \texttt{findings}, \texttt{evidence\_gaps}, \texttt{next\_attempt} (1--5), \texttt{recommendation} = close \\
    \bottomrule
  \end{tabularx}
\end{table}

\paragraph{Audit instructions}
We quote fragments of \sysname's instructions (\ldots\ marks an omission); \cfgH receives the same fragments but lacks the further sentences listed in \cref{tab:configs}.

{\footnotesize\noindent\textit{Every audit:} ``You are the bounded control-audit layer around a code-as-policy robot system. Preserve the native candidate unless there is a concrete, checkable defect. Do not invent observations, simulator state, APIs, or success labels. \ldots{} You may only return the JSON object required by the supplied schema.''\par
\noindent\textit{\planaudit:} ``Audit only the native plan's semantics under the rules below. ACCEPT unchanged, REVISE by returning the complete replacement output\_plan, or REJECT with concrete plan-level defects. A revised plan must remain grounded in the visible native context.''\par
\noindent\textit{\codeaudit:} ``Audit one plan-step body. ACCEPT unchanged, REVISE with final\_body, or REJECT. The final\_body value itself must be bare executable step code: never put a wrapper function or RESULT assignment inside final\_body.''\par
\noindent\textit{\stepaudit:} ``Choose COMMIT only when sandbox success and native per-step status is exactly succeeded; otherwise retry, replan, or abort with a concrete correction.'' \textit{On a plan's last step also:} ``This is the plan's last step: committing it ends the episode and claims that the task is complete. This execution's printed output also holds the checks of the earlier steps, which were replayed before this one. If any of those checks, or this step's own, reports that an intended effect did not occur, do not COMMIT, even when the per-step status is succeeded; retry when the cause lies in this step's own action, replan when it lies in an earlier committed step.''\par
\noindent\textit{\finalaudit:} ``Audit process consistency after this plan attempt's native outcome, and analyse its failure. Every step of the plan was committed, but the environment's own check reports that the task is not complete; the harness will start another plan attempt and pass it your findings. \ldots{} Do not guess hidden simulator state. You cannot change benchmark\_success or request execution. Always return recommendation=close.''\par}

\paragraph{Controller}
\Cref{tab:controller} gives the rules and \cref{alg:commit_supp} the loop of one trial.

\begin{table}[t]
  \centering
  \caption{\textbf{The \controller's rules, read top to bottom after every \steptest.} $a$: the \auditor's proposal; $\kappa$: \mainref{eq:commit}; $r$, $q$: tries left for the step and plans left after this one. $^{\ast}$The coder is also asked to redo the step and check the effects of the earlier ones.}
  \label{tab:controller}
  \small
  \setlength{\tabcolsep}{4pt}
  \begin{tabularx}{\columnwidth}{@{}l>{$}l<{$}>{\raggedright\arraybackslash}X@{}}
    \toprule
    & \text{Predicate} & Decision \\
    \midrule
       & T(s)=1 & trial succeeds \\
    \midrule
    R1 & a=\textsf{commit}\wedge\neg\kappa & \decs{retry}, then R2--R3 \\
    R2 & a=\textsf{retry}\wedge r=0\wedge q>0 & \decs{replan} \\
    R3 & a=\textsf{retry}\wedge r=0\wedge q=0 & \decs{abort} \\
    R4 & a=\textsf{abort}\wedge q>0 & \decs{replan} \\
    R5 & a\in\{\textsf{replan},\textsf{abort}\}\wedge q=0\wedge r>0 & \decs{retry}$^{\ast}$ \\
    R6 & a=\textsf{replan}\wedge r=0\wedge q=0 & \decs{abort} \\
    \midrule
       & \text{otherwise} & $a$ \\
    \bottomrule
  \end{tabularx}
\end{table}

\begin{algorithm}[t]
  \caption{Step-level commit within one trial (\mainref{sec:controller}); $r=J-j$, $q=P-p$; outcome: success.}%
  \label{alg:commit_supp}
  \small
  \algnotext{EndFor}
  \algrenewcommand\algorithmicindent{1.0em}
  \begin{algorithmic}[1]
    \Require task $g$, initial state $s_0$, frozen code, budgets $P$, $J$
    \State open one \auditor conversation $A$; $\mathit{issue}\gets\emptyset$
    \For{$p=1,\dots,P$}
      \State $\pi\gets\mathrm{PlanAudit}_A(\mathrm{Planner}(g,\mathit{issue}))$ \Comment{\gateone}
      \State \textbf{if} $\pi$ rejected \textbf{then next plan}
      \State $C\gets()$; $\mathit{fb}\gets\emptyset$
      \For{$k=1,\dots,|\pi|$}
        \For{$j=1,\dots,J$}
          \State $\tilde c\gets\mathrm{CodeAudit}_A(\mathrm{Coder}(\sigma_k,C,\mathit{fb}))$ \Comment{\gatetwo}
          \State \textbf{if} $\tilde c$ rejected or blocked \textbf{then} $\mathit{fb}\gets$ defects; \textbf{continue}
          \State $(s,e)\gets\mathrm{Exec}(\mathrm{reset}(s_0),C\oplus\tilde c)$ \Comment{\mainref{eq:steptest}}
          \State $v\gets\mathrm{HoldingCheck}(\mathrm{Verifier}(\sigma_k,\tilde c,s,e))$
          \State $a\gets\mathrm{StepAudit}_A(\sigma_k,\tilde c,e,v,r,q)$ \Comment{\gatethree}
          \State $d\gets\mathrm{Controller}(a,\kappa,r,q)$ \Comment{\cref{tab:controller}, R1--R6}
          \State \textbf{if} $d=\textsf{commit}$ \textbf{then} $C\gets C\oplus\tilde c$
          \State \textbf{if} $T(s)=1$ \textbf{then return} outcome
          \State \textbf{if} $d=\textsf{commit}$ \textbf{then} $\mathit{fb}\gets\emptyset$; \textbf{next step}
          \State \textbf{if} $d=\textsf{retry}$ \textbf{then} $\mathit{fb}\gets$ correction
          \State \textbf{if} $d=\textsf{replan}$ \textbf{then} $\mathit{issue}\gets$ correction; \textbf{next plan}
          \State \textbf{if} $d=\textsf{abort}$ \textbf{then return} failure
        \EndFor
        \State \textbf{if} step $k$ uncommitted \textbf{then} $\mathit{issue}\gets$ reason; \textbf{next plan}
      \EndFor
      \State \textbf{if} $p<P$ \textbf{then} $\mathit{issue}\gets\mathrm{FinalAudit}_A(\cdot)$ \Comment{\gatefour}
    \EndFor
    \State \Return failure
  \end{algorithmic}
\end{algorithm}

\section{Between-run refinement and privileged information}
\label{sec:supp_refine}

\Cref{tab:supp_changes} lists the changes to the released code.

\begin{table}[t]
  \centering
  \caption{\textbf{The change register on \libero.} The 194 changes of \sysname relative to the released \rats code, by class, and what \cfgH keeps. The refined library has 71 entries: the 59 released ones, patched, plus 2 guidance entries, 6 new skills and 4 new perception primitives, written by the coding agent.}
  \label{tab:supp_changes}
  \footnotesize
  \setlength{\tabcolsep}{2.5pt}
  \begin{tabularx}{\columnwidth}{@{}>{\raggedright\arraybackslash}p{0.17\columnwidth}c>{\raggedright\arraybackslash}X>{\raggedright\arraybackslash}p{0.17\columnwidth}@{}}
    \toprule
    Class & Items & Contents & In \cfgH \\
    \midrule
    Consistency fixes & 78 & success read only from the \taskcheck; correct initial-state loading; infrastructure faults discarded; an error after the \taskcheck passed counts as a success & all kept (4 rewritten) \\
    Harness wrapper & 49 & rule text (13), skill-interface view (5), control layer (31): audits, \controller, holding check, \finalaudit & all kept (2 rewritten) \\
    Skill revisions & 40 & library and runtime patches (9), new skills and perception interfaces (15), primitive fixes incl.\ the grasp-network frame (12), prompt text (4) & all removed \\
    Undecided & 27 & documentation fact corrections (13), manipulation principles (8), six others & 3 kept, incl.\ grasp-sample reuse \\
    \bottomrule
  \end{tabularx}
\end{table}

\paragraph{Privileged information}
\label{sec:supp_leak}
Leakage is privileged data that reaches the text a run-time model sees or the code the policy executes: object poses and sizes, articulation states, success thresholds and intermediate quantities of the \taskcheck, numbers tuned against ground truth, and facts about a single episode that only the benchmark's files hold.
The final success signal (also per predicate), proprioception, camera data and quantities perceived from them, the task text as the released bridge passes it, the suite name and task type, what a type of task requires in words, and constants set with reference to, but not copied from, asset geometry or the success code are not leakage.
Under this definition we found no leakage in \cfgH or \sysname.

\section{Compared systems and protocol}
\label{sec:supp_config}
\label{sec:supp_pending}
\label{sec:supp_rerun}
\Cref{tab:configs} lists every difference we know of between the five \libero systems, so each comparison of \mainref{sec:ablation} is between complete systems; the settings of \cfgA and \cfgO were fixed before any of their results was seen.

\begin{figure}[t]
  \centering
  \includegraphics[width=0.7\columnwidth]{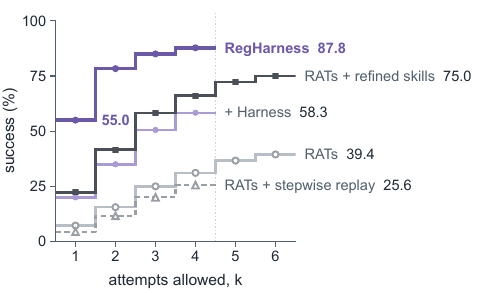}
  \caption{\textbf{\sysname solves 55.0\% of the paired trials with its first plan, more than \rats with its whole budget, and 87.8\% within four; \cfgO reaches 75.0\% with six attempts.} An attempt is one whole program for \rats and \cfgO, one stepwise pass through a plan for \cfgA, and one plan of up to four tries per step for \cfgH and \sysname, so attempts are not equal work; dotted: the four-attempt budget of the last three.}
  \label{fig:results}
\end{figure}
\begin{table*}[t]
  \centering
  \caption{\textbf{Every known difference between the five \libero systems.} ``As X'': as in system X.}
  \label{tab:configs}
  \scriptsize
  \renewcommand{\arraystretch}{1.0}
  \setlength{\tabcolsep}{4pt}
  \begin{tabularx}{\textwidth}{@{}>{\raggedright\arraybackslash}p{1.55cm}>{\hsize=1.10\hsize\raggedright\arraybackslash}X>{\hsize=1.04\hsize\raggedright\arraybackslash}X>{\hsize=.70\hsize\raggedright\arraybackslash}X>{\hsize=1.18\hsize\raggedright\arraybackslash}X>{\hsize=.98\hsize\raggedright\arraybackslash}X@{}}
    \toprule
    & \rats & \cfgA & \cfgO & \cfgH & \sysname \\
    \midrule
    Loop & plan, write the whole program, static check, rehearse, execute, verify, diagnose, retry & \rats' loop with \texttt{-{}-multiturn-reset}: each step tested from a reset, committed steps replayed & as \rats & run-time governance (\mainref{sec:method}) & as \cfgH \\
    Roles & planner, planner verifier, whole-program coder, static checker, per-step verifier, diagnoser, goal verifier, SubAgent & as \rats; the coder writes one step at a time & as \rats & coder writes one step at a time; drops planner verifier, rehearsal, SubAgent, feedback generator and the goal verifier's LLM analysis; diagnoser on ambiguous verdicts only; adds the \auditor & as \cfgH \\
    Budget; time per execution & 6 attempts; 180\,s & 4 attempts, ${\le}4$ tries per step (code default 10); 180\,s per \steptest incl.\ replay & 6 attempts; 180\,s & 4 plans, ${\le}4$ tries per step; 480\,s per \steptest & as \cfgH \\
    Rehearsal & yes, unscored & no & yes & no & no \\
    \taskcheck read & after each scored execution & once, at the end of each attempt; reaching the goal mid-attempt does not end it & as \rats & after every \steptest; the trial ends when it holds & as \cfgH \\
    Commit & none; a retry is a new whole program, prompted with the previous program, the diagnosis and segments marked as working & on verifier \emph{succeeded}, also after an error; forced after the last try & as \rats & \mainref{eq:commit} & as \cfgH \\
    Static checker & whole program, incl.\ three checks only for whole programs (release or placement gated on a verification result, grasp with no later release or placement, missing result report); a block uses an attempt & per step, without them & as \rats & each draft and the composed program, without them; a block uses a try & as \cfgH \\
    Audits & none & none & none & plan, code, step, final & as \cfgH, plus two sentences, on lost instances and on re-measuring a target while holding, and a mention of instance identifiers \\
    Holding check & no & no & no & yes; recognizes only the released way of reading the gripper & yes \\
    Skills & released, 59 entries; its loop writes usage statistics back to the library file & released & refined, 69 entries & released, frozen & refined, 71 entries, frozen, byte-identical in all 300 trials \\
    Primitives and documentation & released (9 exported functions) & as \rats; the step writer sees the first 6,000 characters of the documentation & refined; documentation of 18,546 characters against 6,228 & released & refined \\ %
    Planning rules & none & none & none & 5,844 characters & 10,380 characters \\
    Consistency & fixture reset after every reset & as \rats & as \rats & replay reuses grasp samples; inverse-kinematics state cleared at reset; seeded random numbers; no fixture reset & as \cfgH \\
    Images & video, one timestamped frame per second & as \rats; diagnoser: first and last frames & as \rats & verifier: first and last, or sampled, frames; \stepaudit: first and last agent-view frames; other audits: none & as \cfgH \\
    Initial states; parallel trials & 0--2; 6 or 8 & 0--2; 6 to 10 & 0--2; about 10 & 0--2; 12 & 0--4; 12 \\
    \midrule
    Shared by all & \multicolumn{5}{>{\raggedright\arraybackslash}p{\dimexpr\textwidth-1.55cm-2\tabcolsep\relax}@{}}{Franka arm in \libero at 20\,Hz; agent-view and wrist RGB-D; perception (pointing with Molmo2-8B, segmentation with SAM3, depth back-projection) with simulator camera extrinsics and noise-free depth; Contact-GraspNet and PyRoki; saved initial state $s$ with reset seed $s{+}1$; failure memory off; every role on the provider alias \llm, thinking effort ``high'', no temperature.} \\
    \bottomrule
  \end{tabularx}
\end{table*}

\begin{table*}[t]
  \centering
  \caption{\textbf{Per-task outcomes on the Object, Goal and Spatial suites,} by initial state. \cmark{} success, $\cdot$ failure; for \cfgH and \sysname a failed trial shows its class (\cref{tab:failures}). Object instructions read ``Pick the $\langle$object$\rangle$ and place it in the basket'', Spatial ones ``Pick the akita black bowl $\langle$location$\rangle$ and place it on the plate''.}
  \label{tab:supp_pertask_a}
  \footnotesize
  \renewcommand{\arraystretch}{0.84}
  \setlength{\tabcolsep}{3pt}
  \begin{tabular}{@{}rl*{3}{c}@{\hspace{12pt}}*{3}{c}@{\hspace{12pt}}*{3}{c}@{\hspace{12pt}}*{3}{c}@{\hspace{12pt}}*{5}{c}@{}}
    \toprule
    & & \multicolumn{3}{c}{\rats} & \multicolumn{3}{c}{\shortstack{RATs\,+\\stepwise\\replay}} & \multicolumn{3}{c}{\shortstack{RATs\,+\\refined\\skills}} & \multicolumn{3}{c}{\cfgH} & \multicolumn{5}{c}{\sysname} \\
    \cmidrule(lr){3-5}\cmidrule(lr){6-8}\cmidrule(lr){9-11}\cmidrule(lr){12-14}\cmidrule(l){15-19}
    \# & Instruction & 0 & 1 & 2 & 0 & 1 & 2 & 0 & 1 & 2 & 0 & 1 & 2 & 0 & 1 & 2 & 3 & 4 \\
    \midrule
\multicolumn{19}{@{}l}{\textit{Obj-Pos}}\\
0 & alphabet soup & $\cdot$ & $\cdot$ & \cmark & $\cdot$ & \cmark & \cmark & $\cdot$ & $\cdot$ & \cmark & \cmark & \cmark & \cmark & \cmark & \cmark & G & \cmark & \cmark \\
1 & cream cheese & \cmark & \cmark & \cmark & \cmark & \cmark & \cmark & \cmark & \cmark & \cmark & \cmark & A & \cmark & \cmark & \cmark & \cmark & \cmark & \cmark \\
2 & salad dressing & \cmark & \cmark & \cmark & $\cdot$ & $\cdot$ & $\cdot$ & $\cdot$ & $\cdot$ & \cmark & A & A & A & \cmark & \cmark & \cmark & \cmark & A \\
3 & bbq sauce & \cmark & $\cdot$ & $\cdot$ & \cmark & $\cdot$ & $\cdot$ & \cmark & \cmark & $\cdot$ & \cmark & \cmark & \cmark & \cmark & \cmark & A & A & A \\
4 & ketchup & $\cdot$ & \cmark & $\cdot$ & $\cdot$ & $\cdot$ & \cmark & \cmark & \cmark & \cmark & S & A & \cmark & \cmark & A & \cmark & \cmark & A \\
5 & tomato sauce & $\cdot$ & \cmark & $\cdot$ & $\cdot$ & $\cdot$ & $\cdot$ & \cmark & \cmark & \cmark & \cmark & F & \cmark & \cmark & \cmark & \cmark & \cmark & \cmark \\
6 & butter & \cmark & \cmark & \cmark & \cmark & $\cdot$ & $\cdot$ & \cmark & \cmark & \cmark & \cmark & \cmark & \cmark & \cmark & \cmark & \cmark & \cmark & \cmark \\
7 & milk & $\cdot$ & \cmark & \cmark & $\cdot$ & $\cdot$ & $\cdot$ & \cmark & \cmark & \cmark & \cmark & \cmark & \cmark & \cmark & \cmark & \cmark & \cmark & \cmark \\
8 & chocolate pudding & $\cdot$ & \cmark & $\cdot$ & $\cdot$ & $\cdot$ & $\cdot$ & \cmark & \cmark & \cmark & \cmark & S & S & \cmark & \cmark & \cmark & \cmark & \cmark \\
9 & orange juice & \cmark & \cmark & $\cdot$ & \cmark & \cmark & \cmark & \cmark & \cmark & \cmark & \cmark & \cmark & \cmark & \cmark & \cmark & \cmark & \cmark & \cmark \\
\multicolumn{2}{@{}r}{\textit{Obj-Pos total (of 10)}} & 5 & 8 & 5 & 4 & 3 & 4 & 8 & 8 & 9 & 8 & 5 & 8 & 10 & 9 & 8 & 9 & 7 \\
\addlinespace[2pt]
\multicolumn{19}{@{}l}{\textit{Obj-Task}}\\
0 & cream cheese & \cmark & \cmark & $\cdot$ & \cmark & $\cdot$ & \cmark & \cmark & \cmark & \cmark & S & \cmark & S & \cmark & \cmark & \cmark & \cmark & A \\
1 & alphabet soup & \cmark & $\cdot$ & $\cdot$ & $\cdot$ & $\cdot$ & \cmark & \cmark & \cmark & $\cdot$ & S & A & \cmark & \cmark & \cmark & A & \cmark & \cmark \\
2 & tomato sauce & $\cdot$ & \cmark & $\cdot$ & $\cdot$ & \cmark & \cmark & \cmark & \cmark & $\cdot$ & U & S & A & \cmark & \cmark & A & \cmark & \cmark \\
3 & ketchup & $\cdot$ & $\cdot$ & $\cdot$ & $\cdot$ & \cmark & $\cdot$ & \cmark & \cmark & \cmark & A & \cmark & \cmark & \cmark & \cmark & \cmark & \cmark & A \\
4 & milk & \cmark & \cmark & \cmark & \cmark & $\cdot$ & \cmark & \cmark & \cmark & \cmark & \cmark & \cmark & \cmark & \cmark & \cmark & \cmark & \cmark & \cmark \\
5 & bbq sauce & $\cdot$ & $\cdot$ & \cmark & \cmark & \cmark & \cmark & \cmark & \cmark & \cmark & \cmark & \cmark & \cmark & \cmark & \cmark & \cmark & \cmark & \cmark \\
6 & orange juice & \cmark & \cmark & $\cdot$ & \cmark & \cmark & \cmark & \cmark & \cmark & \cmark & \cmark & \cmark & \cmark & \cmark & \cmark & \cmark & \cmark & \cmark \\
7 & butter & $\cdot$ & $\cdot$ & $\cdot$ & $\cdot$ & $\cdot$ & \cmark & $\cdot$ & \cmark & \cmark & S & A & S & A & A & \cmark & A & A \\
8 & salad dressing & $\cdot$ & $\cdot$ & $\cdot$ & $\cdot$ & $\cdot$ & $\cdot$ & $\cdot$ & $\cdot$ & $\cdot$ & S & A & A & \cmark & \cmark & \cmark & \cmark & \cmark \\
9 & chocolate pudding & $\cdot$ & \cmark & \cmark & $\cdot$ & $\cdot$ & $\cdot$ & \cmark & \cmark & \cmark & \cmark & \cmark & S & \cmark & \cmark & \cmark & \cmark & \cmark \\
\multicolumn{2}{@{}r}{\textit{Obj-Task total (of 10)}} & 4 & 5 & 3 & 4 & 4 & 7 & 8 & 9 & 7 & 4 & 6 & 5 & 9 & 9 & 8 & 9 & 7 \\
\addlinespace[2pt]
\multicolumn{19}{@{}l}{\textit{Goal-Pos}}\\
0 & Open the middle layer of the drawer & $\cdot$ & $\cdot$ & $\cdot$ & $\cdot$ & $\cdot$ & $\cdot$ & $\cdot$ & \cmark & $\cdot$ & U & U & U & \cmark & G & \cmark & \cmark & A \\
1 & Put the bowl on the stove & $\cdot$ & \cmark & \cmark & $\cdot$ & \cmark & $\cdot$ & \cmark & \cmark & \cmark & H1 & F & \cmark & \cmark & \cmark & \cmark & \cmark & \cmark \\
2 & Put the wine bottle on the top of the drawer & $\cdot$ & $\cdot$ & \cmark & \cmark & $\cdot$ & $\cdot$ & \cmark & \cmark & \cmark & \cmark & \cmark & \cmark & \cmark & \cmark & \cmark & \cmark & \cmark \\
3 & Open the top layer of the drawer and put the bowl inside & $\cdot$ & $\cdot$ & $\cdot$ & $\cdot$ & $\cdot$ & $\cdot$ & \cmark & \cmark & $\cdot$ & \cmark & F & U & \cmark & \cmark & \cmark & \cmark & \cmark \\
4 & Put the bowl on the top of the drawer & $\cdot$ & $\cdot$ & \cmark & \cmark & $\cdot$ & $\cdot$ & \cmark & \cmark & \cmark & \cmark & \cmark & \cmark & \cmark & \cmark & \cmark & \cmark & \cmark \\
5 & Push the plate to the front of the stove & $\cdot$ & $\cdot$ & $\cdot$ & $\cdot$ & $\cdot$ & $\cdot$ & $\cdot$ & $\cdot$ & $\cdot$ & U & U & C & C & C & C & C & \cmark \\
6 & Put the cream cheese on the bowl & $\cdot$ & \cmark & \cmark & $\cdot$ & \cmark & $\cdot$ & \cmark & \cmark & \cmark & F & \cmark & \cmark & \cmark & \cmark & \cmark & \cmark & \cmark \\
7 & Turn on the stove & \cmark & $\cdot$ & \cmark & \cmark & \cmark & \cmark & \cmark & $\cdot$ & \cmark & \cmark & \cmark & \cmark & \cmark & \cmark & \cmark & \cmark & \cmark \\
8 & Put the bowl on the plate & $\cdot$ & \cmark & $\cdot$ & $\cdot$ & $\cdot$ & $\cdot$ & \cmark & \cmark & \cmark & \cmark & \cmark & \cmark & \cmark & \cmark & \cmark & \cmark & \cmark \\
9 & Put the wine bottle on the rack & \cmark & \cmark & \cmark & $\cdot$ & $\cdot$ & $\cdot$ & $\cdot$ & \cmark & $\cdot$ & F & \cmark & \cmark & \cmark & \cmark & \cmark & \cmark & \cmark \\
\multicolumn{2}{@{}r}{\textit{Goal-Pos total (of 10)}} & 2 & 4 & 6 & 3 & 3 & 1 & 7 & 8 & 6 & 5 & 6 & 7 & 9 & 8 & 9 & 9 & 9 \\
\addlinespace[2pt]
\multicolumn{19}{@{}l}{\textit{Goal-Task}}\\
0 & open the bottom drawer of the cabinet & $\cdot$ & $\cdot$ & $\cdot$ & $\cdot$ & $\cdot$ & $\cdot$ & $\cdot$ & $\cdot$ & $\cdot$ & D & D & D & D & D & D & D & D \\
1 & Put the plate on the stove & \cmark & $\cdot$ & $\cdot$ & $\cdot$ & $\cdot$ & $\cdot$ & $\cdot$ & \cmark & \cmark & C & \cmark & \cmark & C & C & \cmark & \cmark & C \\
2 & put the wine bottle in the bowl & \cmark & $\cdot$ & \cmark & \cmark & $\cdot$ & $\cdot$ & \cmark & $\cdot$ & \cmark & \cmark & \cmark & \cmark & \cmark & \cmark & \cmark & \cmark & \cmark \\
3 & Open the top layer of the drawer and put the cream cheese inside & \cmark & $\cdot$ & $\cdot$ & \cmark & $\cdot$ & $\cdot$ & \cmark & \cmark & \cmark & S & U & S & \cmark & \cmark & \cmark & \cmark & \cmark \\
4 & Put the plate on the top of the drawer & $\cdot$ & $\cdot$ & $\cdot$ & $\cdot$ & $\cdot$ & $\cdot$ & \cmark & \cmark & \cmark & C & \cmark & C & C & \cmark & C & C & C \\
5 & Push the cream cheese to the front of the stove & $\cdot$ & $\cdot$ & $\cdot$ & \cmark & $\cdot$ & $\cdot$ & $\cdot$ & $\cdot$ & $\cdot$ & U & B & \cmark & \cmark & \cmark & \cmark & \cmark & \cmark \\
6 & put the wine bottle in the bowl & \cmark & $\cdot$ & $\cdot$ & $\cdot$ & $\cdot$ & $\cdot$ & \cmark & $\cdot$ & $\cdot$ & U & \cmark & \cmark & \cmark & \cmark & \cmark & \cmark & \cmark \\
7 & Turn off the stove & \cmark & \cmark & \cmark & \cmark & \cmark & $\cdot$ & \cmark & \cmark & \cmark & \cmark & A & \cmark & \cmark & \cmark & \cmark & \cmark & \cmark \\
8 & Put the wine bottle on the plate & \cmark & \cmark & \cmark & $\cdot$ & \cmark & \cmark & \cmark & $\cdot$ & \cmark & \cmark & \cmark & \cmark & \cmark & \cmark & \cmark & \cmark & \cmark \\
9 & Put the cream cheese on the rack & $\cdot$ & $\cdot$ & \cmark & $\cdot$ & $\cdot$ & \cmark & \cmark & \cmark & \cmark & S & S & \cmark & \cmark & \cmark & \cmark & \cmark & \cmark \\
\multicolumn{2}{@{}r}{\textit{Goal-Task total (of 10)}} & 6 & 2 & 4 & 4 & 2 & 2 & 7 & 5 & 7 & 3 & 5 & 7 & 7 & 8 & 8 & 8 & 7 \\
\addlinespace[2pt]
    \bottomrule
  \end{tabular}
\end{table*}
\begin{table*}[t]
  \centering
  \addtocounter{table}{-1}\renewcommand{\theHtable}{\thetable.cont}%
  \caption{(continued) Spatial suite, and totals over all 60 tasks.}
  \footnotesize
  \renewcommand{\arraystretch}{0.84}
  \setlength{\tabcolsep}{3pt}
  \begin{tabular}{@{}rl*{3}{c}@{\hspace{12pt}}*{3}{c}@{\hspace{12pt}}*{3}{c}@{\hspace{12pt}}*{3}{c}@{\hspace{12pt}}*{5}{c}@{}}
    \toprule
    & & \multicolumn{3}{c}{\rats} & \multicolumn{3}{c}{\shortstack{RATs\,+\\stepwise\\replay}} & \multicolumn{3}{c}{\shortstack{RATs\,+\\refined\\skills}} & \multicolumn{3}{c}{\cfgH} & \multicolumn{5}{c}{\sysname} \\
    \cmidrule(lr){3-5}\cmidrule(lr){6-8}\cmidrule(lr){9-11}\cmidrule(lr){12-14}\cmidrule(l){15-19}
    \# & Instruction & 0 & 1 & 2 & 0 & 1 & 2 & 0 & 1 & 2 & 0 & 1 & 2 & 0 & 1 & 2 & 3 & 4 \\
    \midrule
\multicolumn{19}{@{}l}{\textit{Spa-Pos}}\\
0 & between the plate and the ramekin & \cmark & $\cdot$ & \cmark & $\cdot$ & $\cdot$ & $\cdot$ & \cmark & \cmark & \cmark & \cmark & \cmark & \cmark & \cmark & \cmark & \cmark & \cmark & \cmark \\
1 & next to the ramekin & $\cdot$ & \cmark & $\cdot$ & $\cdot$ & $\cdot$ & $\cdot$ & \cmark & $\cdot$ & \cmark & \cmark & \cmark & \cmark & \cmark & \cmark & \cmark & \cmark & \cmark \\
2 & from table center & $\cdot$ & $\cdot$ & $\cdot$ & $\cdot$ & $\cdot$ & $\cdot$ & \cmark & $\cdot$ & \cmark & S & \cmark & \cmark & \cmark & \cmark & \cmark & \cmark & \cmark \\
3 & on the cookies box & $\cdot$ & $\cdot$ & $\cdot$ & $\cdot$ & $\cdot$ & $\cdot$ & $\cdot$ & \cmark & \cmark & \cmark & \cmark & A & \cmark & \cmark & \cmark & \cmark & \cmark \\
4 & in the top layer of the wooden cabinet & $\cdot$ & $\cdot$ & $\cdot$ & $\cdot$ & $\cdot$ & $\cdot$ & $\cdot$ & $\cdot$ & $\cdot$ & B & B & B & B & B & B & E & B \\
5 & on the ramekin & \cmark & $\cdot$ & $\cdot$ & $\cdot$ & $\cdot$ & $\cdot$ & \cmark & \cmark & \cmark & \cmark & \cmark & F & \cmark & \cmark & \cmark & \cmark & \cmark \\
6 & next to the cookies box & $\cdot$ & \cmark & $\cdot$ & $\cdot$ & $\cdot$ & $\cdot$ & \cmark & \cmark & \cmark & A & H1 & A & \cmark & \cmark & \cmark & \cmark & \cmark \\
7 & on the stove & $\cdot$ & $\cdot$ & \cmark & $\cdot$ & $\cdot$ & $\cdot$ & \cmark & \cmark & $\cdot$ & H2 & A & \cmark & \cmark & \cmark & \cmark & \cmark & \cmark \\
8 & next to the plate & $\cdot$ & $\cdot$ & $\cdot$ & $\cdot$ & $\cdot$ & $\cdot$ & $\cdot$ & \cmark & $\cdot$ & \cmark & \cmark & \cmark & \cmark & \cmark & \cmark & \cmark & \cmark \\
9 & on the wooden cabinet & \cmark & \cmark & \cmark & $\cdot$ & $\cdot$ & \cmark & \cmark & $\cdot$ & \cmark & \cmark & \cmark & \cmark & \cmark & \cmark & \cmark & \cmark & \cmark \\
\multicolumn{2}{@{}r}{\textit{Spa-Pos total (of 10)}} & 3 & 3 & 3 & 0 & 0 & 1 & 7 & 6 & 7 & 6 & 7 & 6 & 9 & 9 & 9 & 9 & 9 \\
\addlinespace[2pt]
\multicolumn{19}{@{}l}{\textit{Spa-Task}}\\
0 & not between the plate and the ramekin & $\cdot$ & $\cdot$ & $\cdot$ & $\cdot$ & $\cdot$ & \cmark & \cmark & \cmark & \cmark & A & \cmark & \cmark & \cmark & \cmark & \cmark & \cmark & \cmark \\
1 & next to the cookie box & $\cdot$ & $\cdot$ & \cmark & $\cdot$ & $\cdot$ & \cmark & \cmark & \cmark & $\cdot$ & S & S & \cmark & \cmark & \cmark & \cmark & \cmark & \cmark \\
2 & next to the plate & $\cdot$ & $\cdot$ & \cmark & \cmark & \cmark & $\cdot$ & \cmark & \cmark & \cmark & \cmark & \cmark & \cmark & \cmark & \cmark & \cmark & \cmark & \cmark \\
3 & on the top of the cabinet & $\cdot$ & $\cdot$ & \cmark & $\cdot$ & $\cdot$ & $\cdot$ & \cmark & \cmark & \cmark & F & \cmark & \cmark & \cmark & \cmark & \cmark & \cmark & \cmark \\
4 & on the top of the wooden cabinet & $\cdot$ & $\cdot$ & $\cdot$ & $\cdot$ & $\cdot$ & $\cdot$ & \cmark & \cmark & \cmark & U & S & U & \cmark & \cmark & \cmark & \cmark & \cmark \\
5 & on the cookie box & $\cdot$ & $\cdot$ & \cmark & $\cdot$ & $\cdot$ & $\cdot$ & \cmark & $\cdot$ & \cmark & \cmark & F & \cmark & \cmark & \cmark & \cmark & \cmark & \cmark \\
6 & on the stove & $\cdot$ & \cmark & $\cdot$ & $\cdot$ & $\cdot$ & $\cdot$ & \cmark & \cmark & \cmark & \cmark & H2 & H2 & \cmark & \cmark & F & \cmark & \cmark \\
7 & on the top of the cabinet & $\cdot$ & $\cdot$ & $\cdot$ & $\cdot$ & $\cdot$ & $\cdot$ & \cmark & $\cdot$ & \cmark & \cmark & F & \cmark & \cmark & \cmark & \cmark & \cmark & \cmark \\
8 & next to the ramekin & \cmark & \cmark & $\cdot$ & $\cdot$ & $\cdot$ & $\cdot$ & \cmark & \cmark & $\cdot$ & \cmark & F & \cmark & \cmark & \cmark & \cmark & \cmark & \cmark \\
9 & on the stove & \cmark & $\cdot$ & $\cdot$ & $\cdot$ & $\cdot$ & $\cdot$ & \cmark & \cmark & \cmark & F & \cmark & \cmark & \cmark & \cmark & \cmark & A & \cmark \\
\multicolumn{2}{@{}r}{\textit{Spa-Task total (of 10)}} & 2 & 2 & 4 & 1 & 1 & 2 & 10 & 8 & 8 & 5 & 4 & 8 & 10 & 10 & 9 & 9 & 10 \\
\addlinespace[2pt]
\midrule
\multicolumn{2}{@{}r}{\textit{All 60 tasks}} & 22 & 24 & 25 & 16 & 13 & 17 & 47 & 44 & 44 & 31 & 33 & 41 & 54 & 53 & 51 & 53 & 49 \\
    \bottomrule
  \end{tabular}
\end{table*}

\paragraph{Running \rats}
Our \rats baseline is the released code~\cite{zhang2026rats} with five files changed and a local proxy added.
It uses the full perception API with a Contact-GraspNet grasp sampler, because under the default evaluation interface 36 of the 45 released skills in use cannot be called.

\paragraph{Rehearsal}
\label{sec:supp_rehearsal}
The released configuration runs every program that passes the static checker once from the reset scene before its scored execution, a stage the RATs paper does not describe; both runs execute the same program string from the same reset scene and seed, and the rehearsal's \taskcheck is not used.
It was true at the end of 77 of the 782 rehearsals, in 65 of the 180 trials.
Over the 741 pairs of a rehearsal and the scored execution of the same program (the other 41 rehearsals raised an error), the outcome differs in 76 (exact McNemar $p=0.57$).

\paragraph{\cfgA}
\label{sec:supp_stepwise}
In the stepwise mode, the released per-step verifier indexes a single step as the first. \cfgA corrects this indexing error in the per-step verifier and changes no other code.
Executions in goal tasks reach the 180\,s limit more often when more trials share the machine, which may lower its result.

\paragraph{\cfgO}
It carries 51 of the 64 changes that \cfgH removes from \sysname, 43 verbatim and 8 rewritten; one more, the pair of guidance entries, never reaches the model in either framework, and twelve exist only in the planning rules, audit instructions or modified planner that \rats lacks.
\rats' static checker rejects programs that gate a release or placement on a verification result, which the holding-verification skills' documentation encourages, and each rejection, for whatever reason, uses up an attempt (139 of this system's 680 checked programs); the SubAgent cannot call the new skills.

\paragraph{Voided runs}
Each trial is scored once: a process stopped by an infrastructure or protocol fault was discarded, and the trial was rerun from the start with a full budget.
On \libero this voided 13 of the 493 processes of the two harness systems, none after the \taskcheck had passed; 10 reruns succeeded, two of them with more than four plans over both processes.
No \rats process was voided, and interrupted \cfgA and \cfgO trials were rerun the same way.
On \molmo six trials were rerun and all six succeeded, one of them after an out-of-memory kill caused by the policy's own code, which we counted as an infrastructure fault; \molmo trials run with an earlier set of switches are not reported.

\paragraph{Statistics}
Intervals for rates and paired differences come from a bootstrap that resamples the 60 tasks with all initial states of a task together; we report the 2.5th and 97.5th percentiles.
Paired tests are two-sided exact McNemar tests on the discordant trials.

\section{Additional \libero results}
\label{sec:supp_pertask}
\label{sec:supp_results}
\label{sec:supp_state2}

\Cref{tab:supp_pertask_a} gives the outcome of every trial and, for \cfgH and \sysname, the class of every failed one; \cref{tab:supp_paired} gives the paired comparisons of \mainref{sec:ablation}, and \cref{fig:results} shows when the successes arrive.

\begin{table*}[t]
  \centering
  \caption{\textbf{Paired comparisons on the 180 paired trials}: discordant trials, exact two-sided McNemar $p$, and the difference in points with its task-clustered bootstrap 95\% interval (\cref{sec:supp_rerun}).}
  \label{tab:supp_paired}
  \footnotesize
  \setlength{\tabcolsep}{5pt}
  \begin{tabular}{@{}lccccc@{}}
    \toprule
    First $\rightarrow$ second & Only 2nd & Only 1st & $p$ & Diff. & 95\% interval \\
    \midrule
    \multicolumn{6}{@{}l}{\textit{Run-time governance}} \\
    \quad \rats $\rightarrow$ \cfgH & 55 & 21 & $1.2{\times}10^{-4}$ & $+18.9$ & [$+9.4$, $+28.3$] \\
    \quad \cfgO $\rightarrow$ \sysname & 31 & 8 & $2.9{\times}10^{-4}$ & $+12.8$ & [$+5.6$, $+20.6$] \\
    \multicolumn{6}{@{}l}{\textit{Refined skills}} \\
    \quad \rats $\rightarrow$ \cfgO & 74 & 10 & $3.3{\times}10^{-13}$ & $+35.6$ & [$+26.1$, $+45.0$] \\
    \quad \cfgH $\rightarrow$ \sysname & 57 & 4 & $4.9{\times}10^{-13}$ & $+29.4$ & [$+21.1$, $+38.3$] \\
    \multicolumn{6}{@{}l}{\textit{\rats' stepwise mode}} \\
    \quad \rats $\rightarrow$ \cfgA & 18 & 43 & 0.0019 & $-13.9$ & [$-22.8$, $-5.0$] \\
    \quad \cfgA $\rightarrow$ \cfgH & 69 & 10 & $5.5{\times}10^{-12}$ & $+32.8$ & [$+22.8$, $+42.8$] \\
    \multicolumn{6}{@{}l}{\textit{Other pairs}} \\
    \quad \cfgH $\rightarrow$ \cfgO & 50 & 20 & $4.4{\times}10^{-4}$ & $+16.7$ & [$+6.7$, $+26.7$] \\
    \quad \rats $\rightarrow$ \sysname & 90 & 3 & $2.7{\times}10^{-23}$ & $+48.3$ & [$+40.0$, $+56.7$] \\
    \bottomrule
  \end{tabular}
\end{table*}

\paragraph{How \rats retries}
For the 574 retries whose previous attempt was executed and diagnosed with a failed step, we split both programs at the step markers into a preamble and one segment per step, ignore comments and indentation, and count a segment as changed when its similarity to the previous version of the same step is below 0.9; in 493 (86\%) a segment other than the failed step's changes.

\subsection{Scene consistency under reset}
\label{sec:supp_scene}
\begin{figure}[t]
  \centering
  \begin{subfigure}[b]{0.40\columnwidth}\includegraphics[width=\linewidth]{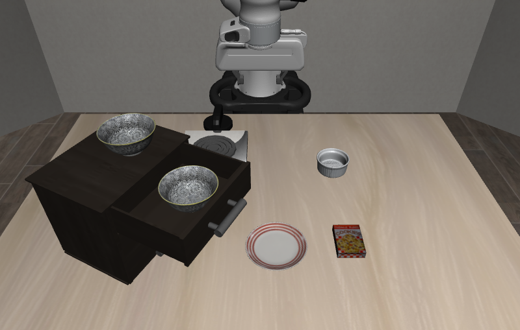}\caption{Spa-Pos 4, first step test}\end{subfigure}\hspace{6pt}
  \begin{subfigure}[b]{0.40\columnwidth}\includegraphics[width=\linewidth]{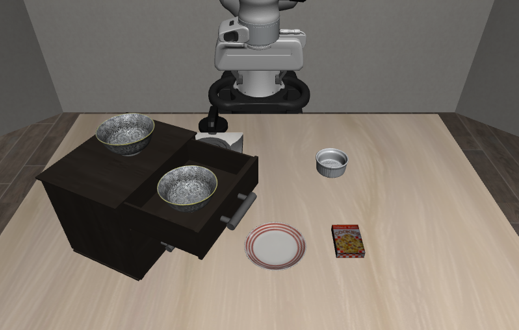}\caption{Spa-Pos 4, second plan}\end{subfigure}\\[2pt]
  \begin{subfigure}[b]{0.40\columnwidth}\includegraphics[width=\linewidth]{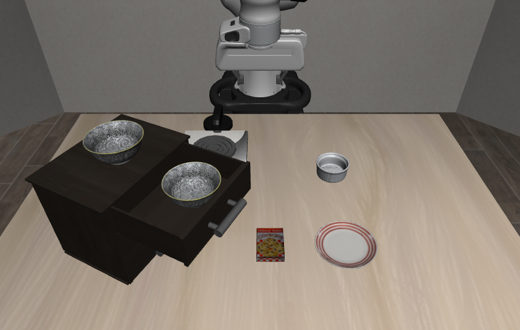}\caption{Spa-Task 4, first step test}\end{subfigure}\hspace{6pt}
  \begin{subfigure}[b]{0.40\columnwidth}\includegraphics[width=\linewidth]{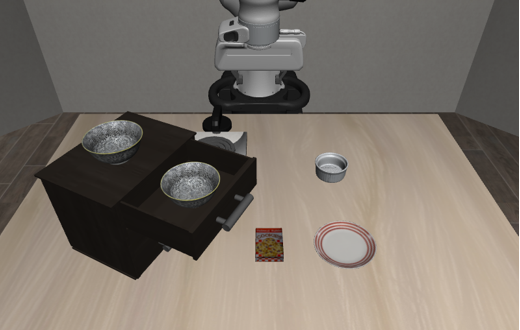}\caption{Spa-Task 4, fourth plan}\end{subfigure}
  \caption{\textbf{In two tasks the cabinet settles differently after each reset.} Each row: the first frames of two step tests with no committed steps in one \cfgH trial. The cabinet, with the bowls on and in it, is displaced; the plate and the box are not.}
  \label{fig:supp_scene}
\end{figure}

Replay re-executes committed code without restoring the simulator state (\mainref{sec:replay}), so we compared rendered frames pixel by pixel.
\rats, \cfgH and \sysname start a trial from the same scene (states 0 and 1), and within a trial of either harness system the scene is the same after every reset (fewer than 0.4\% of the pixels differ), apart from the arm's starting pose in two trials and from Spa-Pos~4 and Spa-Task~4, whose cabinet, with an open drawer, settles in a different pose after every reset (\cref{fig:supp_scene}) for a reason we have not found.
These two tasks are 6 of the 180 paired trials, of which \sysname solves three and \rats and \cfgH none.

\subsection{Inside a trial}
\label{sec:supp_usage}
\label{sec:supp_metrics}
\Cref{tab:usage} gives the counts behind \mainref{sec:metrics}.
A step test is rejected when it leads to a retry, a new plan or an abort; the arm has moved when at least one motion command of the step under test, not of a replayed step, advanced the simulation by at least one frame.

\begin{table}[t]
  \centering
  \caption{\textbf{How often each mechanism of \rtgov acts}, on the 180 paired trials and, in the last column, all 300 trials of \sysname. ``Passed'' refers to the per-step verifier and ``proposed'' to the \stepaudit.}
  \label{tab:usage}
  \label{tab:supp_metrics}
  \footnotesize
  \setlength{\tabcolsep}{1.5pt}
  \begin{tabular}{@{}lrrr@{}}
    \toprule
    & \cfgH & \multicolumn{2}{c}{\sysname} \\
    \cmidrule(lr){2-2}\cmidrule(l){3-4}
    Trials & 180 & 180 & 300 \\
    \midrule
    Plans audited & 530 & 327 & 544 \\
    \quad rewritten & 52.8\% & 67.6\% & 67.3\% \\
    Drafts audited & 3,207 & 2,660 & 4,460 \\
    \quad rewritten & 56.5\% & 48.5\% & 48.5\% \\
    Step tests (per trial) & 3,185 (17.7) & 2,648 (14.7) & 4,440 (14.8) \\
    \quad committed & 1,620 & 1,973 & 3,265 \\
    \quad rejected & 1,547 & 642 & 1,119 \\
    \quad rejected, arm had moved & 1,157 & 264 & 452 \\
    Solved trials with such a test & 84/105 & 55/158 & 87/260 \\
    Passed, commit not proposed & 91/1,712 & 84/2,057 & 135/3,400 \\
    Holding check overturned a pass & 19 & 0 & 3 \\
    Proposals changed by \controller & 56 & 33 & 62 \\
    Closed by \taskcheck, no commit & 18 & 33 & 56 \\
    Final audits (trials; solved later) & 46 (44; 29) & 19 (15; 11) & 30 (25; 16) \\
    \midrule
    Replayed steps per step test & 1.8 & 3.1 & 3.1 \\
    Replayed share of step segments & 63.9\% & 75.8\% & 75.7\% \\
    Prefix-replay aborts (trials) & 30 (16) & 4 (4) & 9 (6) \\
    False-commit plans (trials) & 54/141 (50) & 22/147 (17) & 34/238 (27) \\
    \bottomrule
  \end{tabular}
\end{table}

\paragraph{Prefix-replay aborts}
A step test aborts in its prefix when replaying the committed steps raises an error, so the step under test never runs.
The count is a lower bound, since a replay that ends with a different outcome is not counted, and the lower rate of \sysname does not show more faithful replays, because the skill libraries differ in how many checks can raise.

\paragraph{False-commit proxy}
A plan in which every step that ran was committed while the \taskcheck failed contains at least one false commit.
We count such plans among all plans whose executed steps were all committed; the count is a lower bound and does not say which step was wrong.

\section{Failures and cost}
\label{sec:supp_failures}
\label{sec:supp_cost}

\paragraph{Classification}
Each failed trial receives one class (\cref{tab:failures}; per trial in \cref{tab:supp_pertask_a}): the first error, in time order, that no later plan overcomes, or U when no single error dominates or can be attributed.
An LLM agent assigns the class from the trial's logs and, for some trials, from step-audit frames, and the authors checked every label.
Classes S and H1--H4 were considered for 13 of \sysname's 40 failures only, and the \cfgH failures on state~0 were classified from text, so some of them classed S may be misidentifications.

\begin{table}[t]
  \centering
  \caption{\textbf{Failure classes, one per failed trial.} --: none among the 13 \sysname failures checked for these classes; the other 27 were not checked. No trial is classed H3 (budget used up on a feasible task) or H4 (other error caused by \rtgov).}
  \label{tab:failures}
  \footnotesize
  \setlength{\tabcolsep}{2.3pt}
  \begin{tabular}{@{}l>{\raggedright\arraybackslash}p{0.55\columnwidth}ccc@{}}
    \toprule
    & & \cfgH & \multicolumn{2}{c}{\sysname} \\
    \cmidrule(lr){3-3}\cmidrule(l){4-5}
    & Class & 180 & 180 & 300 \\
    \midrule
    A & Misidentified object or part & 17 & 6 & 16 \\
    B & Instruction or target region misread & 4 & 3 & 4 \\
    C & Flat plate not grasped or pushed & 4 & 7 & 11 \\
    D & Low handle out of reach & 3 & 3 & 5 \\
    E & Holding-verification false negative taken as a stop & 0 & 0 & 1 \\
    F & Grasped but placed off target & 11 & 1 & 1 \\
    G & One-off physical or planning slip & 0 & 2 & 2 \\
    S & Released skill fails to grasp a non-flat object & 19 & -- & -- \\
    H1 & \auditor's rewrite introduced the error & 2 & -- & -- \\
    H2 & A correct step was not committed & 3 & -- & -- \\
    U & Undetermined & 12 & 0 & 0 \\
    \midrule
    & Failed trials & 75 & 22 & 40 \\
    \bottomrule
  \end{tabular}
\end{table}

\begin{table}[t]
  \centering
  \caption{\textbf{Estimated model cost (CNY) and wall-clock time per trial on the 180 paired \libero trials.} Parentheses: the \auditor's calls, the cached share of the input, medians. Costs include discarded processes; times depend on how many trials ran in parallel (\cref{tab:configs}).}
  \label{tab:supp_cost}
  \footnotesize
  \setlength{\tabcolsep}{2pt}
  \begin{tabular}{@{}lrrr@{}}
    \toprule
    & \rats & \cfgH & \sysname \\
    \midrule
    Calls per trial (\auditor) & 57.3 & 79.5 (39.8) & 65.6 (33.0) \\
    Input tokens, M (cached) & 0.54 (23.2\%) & 5.75 (89.5\%) & 5.56 (87.4\%) \\
    Output tokens, M & 0.274 & 0.386 & 0.341 \\
    Cost per trial & 1.51 (1.51) & 2.25 (2.25) & 2.16 (1.42) \\
    Cost per success & 3.83 & 3.85 & 2.46 \\
    Minutes per trial & 50 (46) & 70 (72) & 66 (43) \\
    \bottomrule
  \end{tabular}
\end{table}

\paragraph{Failures caused by run-time governance}
Five \cfgH failures fall into the H classes.
Spa-Pos~5 on state~2 is a borderline case that we count as F; counted as H2, it would make them six.
Two of \sysname's other 27 failures, both classed A, may belong to H2.
Of the nine undetermined \cfgH trials on states~0 and~1, seven mostly stop at a pose check that cannot pass, which traces to the documentation and to audits that asked for it.

\paragraph{Cost}
\phantomsection\label{sec:supp_costest}
Costs are estimated from the logged tokens of every model call of a trial, the \auditor's and those of discarded processes included, at the provider's off-peak list prices (CNY 0.02, 1.0 and 4.0 per million cached input, uncached input and output tokens).
\Cref{tab:supp_cost} gives the cost per trial; \sysname's 300 trials cost CNY 663.27, and the 180 trials of \cfgA and of \cfgO 583.46 and 186.53.

\paragraph{Coding-agent usage}
Between-run refinement as a whole (diagnosing failures; designing, implementing, reviewing and testing changes, dropped ones included; leakage audits, equivalence checks and freezing) took the coding agent, Claude Opus~5.5, an estimated 9.9 billion input tokens on \libero and 2.8 billion on \molmo, over 97\% of them prompt-cache reads, and at least 33.6 and 13.7 million output tokens.
Reports, deployment and run tools, protocol and score analyses are not counted; on \molmo only sub-agents are.

\flushcolsend
\clearpage
\raggedcolsend
\twocolumn[{\begin{minipage}{\textwidth}\captionsetup{type=table}
  \centering
  \caption{\textbf{Outcome of every \molmo task in the five repeats (columns 0--4).} \cmark{} success; \xmark{} failure; -- counted as a failure (the instruction names no placement target); $^{\mathrm{e}}$ error after the \taskcheck had passed (a success); $^{\mathrm{r}}$ rerun after an infrastructure fault; $^{*}$ reaches the agent rewritten (\cref{sec:supp_molmo}). By category: close 46/50, open 42/50, pick 41/50 and pick-and-place 15/50; the \rats paper reports 73.0, 20.0, 37.0 and 22.0\% (400 trials, not run by us).}
  \label{tab:supp_molmo_tasks}
  \small
  \renewcommand{\arraystretch}{1.08}
  \setlength{\tabcolsep}{2pt}
  \begin{minipage}[t]{0.40\textwidth}\centering
  \begin{tabular}[t]{@{}r@{\hspace{4pt}}p{0.37\linewidth}*{5}{c}r@{}}
    \toprule
    & Instruction & 0 & 1 & 2 & 3 & 4 & \\
    \midrule
    \multicolumn{8}{@{}l}{\textit{Close}} \\
    0 & close the table. & \cmark & \cmark$^{\mathrm{r}}$ & \cmark & \cmark & \cmark & 5/5 \\
    1 & close the dresser. & \xmark & \cmark & \cmark & \cmark & \cmark$^{\mathrm{e}}$ & 4/5 \\
    2 & close the table. & \cmark & \cmark & \cmark & \cmark & \cmark & 5/5 \\
    3 & close the dresser. & \cmark & \cmark & \cmark & \cmark & \cmark & 5/5 \\
    4 & Open the drawer$^{*}$ & \cmark & \cmark & \cmark & \cmark & \cmark & 5/5 \\
    5 & close the drawer. & \cmark & \cmark$^{\mathrm{e},\mathrm{r}}$ & \cmark & \cmark & \cmark & 5/5 \\
    6 & Open the dark wood dresser with drawers$^{*}$ & \cmark & \cmark & \cmark & \cmark & \cmark & 5/5 \\
    7 & close the table. & \cmark$^{\mathrm{e}}$ & \xmark & \cmark & \cmark$^{\mathrm{e}}$ & \cmark & 4/5 \\
    8 & close the table. & \cmark & \cmark & \xmark & \cmark & \xmark & 3/5 \\
    9 & Open the dark floral dresser$^{*}$ & \cmark$^{\mathrm{r}}$ & \cmark & \cmark & \cmark & \cmark & 5/5 \\
    \addlinespace[2pt]
    \multicolumn{8}{@{}l}{\textit{Open}} \\
    10 & Open the cabinet & \xmark & \cmark & \cmark & \cmark & \cmark & 4/5 \\
    11 & open the cabinet. & \cmark & \cmark & \cmark & \cmark & \cmark & 5/5 \\
    12 & open the dresser. & \cmark & \cmark & \cmark & \cmark & \cmark & 5/5 \\
    13 & open the cabinet. & \cmark & \cmark$^{\mathrm{r}}$ & \cmark & \cmark & \cmark & 5/5 \\
    14 & open the cabinet. & \cmark & \cmark$^{\mathrm{r}}$ & \cmark & \cmark & \cmark & 5/5 \\
    15 & open the table. & \cmark & \cmark & \cmark & \cmark & \cmark & 5/5 \\
    16 & open the drawer. & \cmark & \xmark & \cmark & \xmark & \xmark & 2/5 \\
    17 & Open the drawer & \cmark & \cmark & \cmark & \cmark & \xmark & 4/5 \\
    18 & open the table. & \xmark & \cmark & \cmark & \xmark & \xmark & 2/5 \\
    19 & open the cabinet. & \cmark & \cmark & \cmark & \cmark & \cmark & 5/5 \\
    \bottomrule
  \end{tabular}
  \end{minipage}
  \hfill
  \begin{minipage}[t]{0.59\textwidth}\centering
  \begin{tabular}[t]{@{}r@{\hspace{4pt}}p{0.56\linewidth}*{5}{c}r@{}}
    \toprule
    & Instruction & 0 & 1 & 2 & 3 & 4 & \\
    \midrule
    \multicolumn{8}{@{}l}{\textit{Pick}} \\
    20 & Pick up the cylindrical dark fabric nhl pencilcase & \cmark & \xmark & \xmark & \xmark & \cmark & 2/5 \\
    21 & Pick up the vintage yellow track shoe with spikes & \cmark & \cmark & \cmark & \cmark & \cmark$^{\mathrm{e}}$ & 5/5 \\
    22 & Pick up the black grenade with red panels & \cmark & \cmark$^{\mathrm{e}}$ & \cmark & \cmark$^{\mathrm{e}}$ & \cmark & 5/5 \\
    23 & Pick up the curved green handle knife with black blade & \cmark & \cmark & \cmark & \cmark & \cmark & 5/5 \\
    24 & pick up the colorful mosaic. & \cmark & \cmark & \cmark & \xmark & \xmark & 3/5 \\
    25 & pick up the black lens. & \cmark & \cmark & \cmark & \cmark & \cmark & 5/5 \\
    26 & Pick up the blocky wooden rifle wrapped in green & \cmark$^{\mathrm{r}}$ & \cmark$^{\mathrm{e}}$ & \cmark & \cmark & \cmark & 5/5 \\
    27 & Pick up the smooth golden bell with knob & \cmark & \cmark$^{\mathrm{e}}$ & \cmark$^{\mathrm{e}}$ & \cmark$^{\mathrm{e}}$ & \cmark & 5/5 \\
    28 & pick up the packet. & \xmark & \cmark & \xmark & \cmark & \xmark & 2/5 \\
    29 & Pick up the black high heel & \cmark & \cmark & \xmark & \cmark & \cmark & 4/5 \\
    \addlinespace[2pt]
    \multicolumn{8}{@{}l}{\textit{Pick-and-place (P\&P)}} \\
    30 & Pick up the electronic equipment and place it in or on the gray bowl & \xmark & \cmark & \cmark & \cmark & \xmark & 3/5 \\
    31 & Pick up the spanner and place it in or on the white bowl blue & \cmark & \cmark & \cmark$^{\mathrm{e}}$ & \cmark & \cmark & 5/5 \\
    32 & pick up the clock. & \xmark & -- & -- & -- & -- & 0/5 \\
    33 & pick up the heart-shaped pink key. & \xmark & -- & -- & -- & -- & 0/5 \\
    34 & pick up the gray shackle. & \xmark & -- & -- & -- & -- & 0/5 \\
    35 & Pick up the red futuristic blaster and place it in or on the gray bowl & \xmark & \xmark & \cmark & \cmark & \xmark & 2/5 \\
    36 & Pick up the black digital camcorder and place it in or on the simple light brown ceramic bowl & \cmark & \cmark & \cmark & \cmark & \cmark & 5/5 \\
    37 & pick up the origami paper crane. & \xmark & -- & -- & -- & -- & 0/5 \\
    38 & pick up the red onion bulb. & \xmark & -- & -- & -- & -- & 0/5 \\
    39 & pick up the green sponge. & \xmark & -- & -- & -- & -- & 0/5 \\
    \midrule
    \multicolumn{2}{@{}l}{Successes per repeat (of 40)} & 28 & 30 & 30 & 30 & 26 & 144/200 \\
    \bottomrule
  \end{tabular}
  \end{minipage}
\end{minipage}
\par\vspace{6pt}}]
\section{\molmo and the real robot}
\label{sec:supp_molmo}

\subsection{\molmo}
We use the 40-episode subset of the \rats paper~\cite{zhang2026rats}, ten test-split episodes each of opening, closing, picking and pick-and-place, and score only the simulator's \texttt{judge\_success()} predicate, as that paper does.
The five repeats rebuild the same 40 scenes; they are repetitions, not different initial states.
All repeats ran with the same code, switches and prompts, with two plans per trial, up to four tries per step and \rats' own retry loop disabled (results in \cref{tab:supp_molmo_tasks}).

\paragraph{Three rewritten instructions}
For close tasks 4, 6 and 9 the upstream generator writes ``Open the \ldots''; the bridge released with \rats replaces ``Open'' with ``Close'' before the agent sees the instruction, and we keep it as released.
All 15 of these trials succeed; we treat the rewrite as part of the \rats protocol and keep the score.

\flushcolsend
\clearpage
\raggedcolsend
\twocolumn[{\begin{minipage}{\textwidth}\captionsetup{type=table}
  \centering
  \caption{\textbf{Real-robot tasks.} Success is judged once the objects have been still for 3\,s; an object still in the gripper is a failure.}
  \label{tab:supp_realrobot_tasks}
  \small%
  \begin{tabularx}{\textwidth}{@{}l>{\raggedright\arraybackslash}p{0.33\textwidth}>{\raggedright\arraybackslash}X@{}}
    \toprule
    Task & Instruction & Success criterion \\
    \midrule
    1 & ``Pick the ketchup and place it in the basket'' & The bottle does not touch the table and is supported by the basket (its rim included), with its midpoint inside the rim seen from above; the distractor bottle is not in the basket; the basket does not tip over \\
    2 & ``Put the cream cheese on the bowl'' & The package is still and touches only the bowl, not the table; the bowl is upright (looser than in simulation, since the real package may only fit across the rim) \\
    3 & ``pick up the cube on the platform, move it off the platform, and exchange it with the cube initially placed below the platform'' & The red cube rests on top of the box; the green cube rests on the table and does not touch the top of the box; the box neither tips over nor moves more than 5\,cm \\
    \bottomrule
  \end{tabularx}
\end{minipage}
\par\vspace{14pt}
  \begin{minipage}{\textwidth}\captionsetup{type=table}
\centering
\small
\setlength{\tabcolsep}{6pt}
\caption{\textbf{Concurrent harnesses, compared on what each paper reports.} \cmark{} reported; $\circ$ in part; \xmark{} not reported.}
\label{tab:concurrent}
\begin{tabular}{@{}lcccccc@{}}
\toprule
 & Run-time & Between-run & Automatic & Shared skills, & Matched & Real \\
 & commit gate & skill/harness edits & acceptance & no VLA or demos & attribution & robot \\
\midrule
Auriga~\cite{wang2026auriga}               & $\circ^{a}$ & $\circ^{b}$ & \xmark & $\circ^{c}$ & $\circ^{d}$ & \xmark \\
RegenHarness~\cite{wang2026regenharness}   & \cmark      & $\circ^{e}$ & \xmark & $\circ^{f}$ & \xmark      & \cmark \\
DynaHarness~\cite{deng2026dynaharness}     & $\circ^{a}$ & \cmark      & \cmark & \xmark      & $\circ^{d}$ & \cmark \\
VISTA~\cite{zhang2026vista}                & $\circ^{a}$ & \cmark      & \cmark & \xmark      & $\circ^{d}$ & \xmark \\
AMP1~\cite{chen2026amp}                    & $\circ^{a}$ & $\circ^{g}$ & \xmark & $\circ^{h}$ & $\circ^{d}$ & \cmark \\
\midrule
\sysname (ours)                            & \cmark      & \cmark      & \xmark & \cmark      & \cmark      & \cmark \\
\bottomrule
\end{tabular}
\par\vspace{2pt}
\parbox{\linewidth}{\footnotesize
Commit gate: fixed rules decide whether an executed step is committed. Automatic acceptance: an edit is kept only after a fixed comparison or held-out test. Matched attribution: gains split between the two levels on paired trials.
$^{a}$Motion checks, no commit decision.
$^{b}$Who revised it is not stated.
$^{c}$No library shared across tasks.
$^{d}$Component ablations.
$^{e}$Protocol only, no results.
$^{f}$Quadruped skills.
$^{g}$Agent-written memory.
$^{h}$Only AMP1+ carries a code library across tasks.}
\end{minipage}
\par\vspace{6pt}}]
\subsection{Real robot}
\label{sec:supp_realrobot}

Both systems run on a 7-DoF Kinova Gen3 arm with a ChangingTek CTAG2F90C two-finger gripper and one fixed Orbbec Gemini 336L RGB-D camera, without a wrist camera (\mainref{fig:realrobot}).
Both use \llm, observe only the camera stream and the joint and gripper readings, and reach the hardware through the same perception and motion services and the same bridge, which rejects commands outside its workspace and motion limits.

\paragraph{Manual reset}
Each step normally continues from the state left by the previous step. If a problem arises, the operator can reset the scene by hand and retry within the same step submission. A failed step can be retried up to three times. After each trial and before a second attempt, the operator resets the scene and the arm returns to its initial configuration.

\Needspace{5\baselineskip}
\paragraph{Protocol}
Each system ran 15 trials per task (tasks in \cref{tab:supp_realrobot_tasks}, results in \mainref{tab:realrobot}), each of at most two attempts, with its skill library fixed across trials, an empty cross-trial memory at the start of every trial and the diagnosis of the first attempt passed to the second.
The operator judges success on site by the criteria of \cref{tab:supp_realrobot_tasks}; this verdict decides whether a second attempt follows and is the only signal returned to the system.
The frames of \mainref{fig:realrobot}(b) come from a run recorded while we set up the tasks, not from the evaluated trials; it replayed, with two distractors added, a program \rats wrote on site.
In the real-robot runs we observed that success depended strongly on where the objects were initially placed and on the randomness of the language model's strategy in a run; more real-robot analysis is left to future work, since here the robot serves only to verify the feasibility of our method.

\section{Comparison with concurrent work}
\label{sec:supp_concurrent}
\Cref{tab:concurrent} compares \sysname with five of the concurrent harnesses of \mainref{sec:related}, on what each paper reports.

\flushcolsend
\clearpage
\raggedcolsend
\end{document}